\documentclass{article}

\usepackage[accepted]{icml2026}
\usepackage{times}

\usepackage{microtype}
\usepackage{graphicx}
\usepackage{subcaption}
\usepackage{booktabs}
\usepackage[table]{xcolor}
\definecolor{lightgray}{gray}{0.9}

\usepackage{amsmath}
\usepackage{amssymb}
\usepackage{mathtools}
\usepackage{amsthm}
\usepackage{amsfonts}

\usepackage{algorithm}
\usepackage{algorithmic}
\usepackage{multirow}
\usepackage[htt]{hyphenat}
\usepackage{stfloats}
\usepackage{url}
\usepackage{hyperref}
\usepackage[capitalize,noabbrev]{cleveref}

\usepackage{amsmath,amsfonts,bm}

\def\eqref#1{equation~\ref{#1}}
\def\1{\bm{1}}

\DeclareMathAlphabet{\mathsfit}{\encodingdefault}{\sfdefault}{m}{sl}
\SetMathAlphabet{\mathsfit}{bold}{\encodingdefault}{\sfdefault}{bx}{n}

\theoremstyle{plain}

\theoremstyle{definition}

\theoremstyle{remark}

\icmltitlerunning{Adaptive Residual-Update Steering for Hallucination Mitigation}

\begin{document}

\twocolumn[
\icmltitle{Adaptive Residual-Update Steering for Low-Overhead Hallucination\\
Mitigation in Large Vision Language Models}
\icmlsetsymbol{equal}{*}

\begin{icmlauthorlist}
\icmlauthor{Zhengtao Zou}{aalto}
\icmlauthor{Ya Gao}{aalto}
\icmlauthor{Jiarui Guan}{aalto}
\icmlauthor{Bin Li}{siat,equal}
\icmlauthor{Pekka Marttinen}{aalto,equal}
\end{icmlauthorlist}

\icmlaffiliation{aalto}{Aalto University, Espoo, Finland}
\icmlaffiliation{siat}{Shenzhen Institutes of Advanced Technology, Chinese Academy of Sciences, Shenzhen, China}

\icmlcorrespondingauthor{Bin Li}{b.li2@siat.ac.cn}
\icmlcorrespondingauthor{Pekka Marttinen}{pekka.marttinen@aalto.fi}

\icmlkeywords{Machine Learning, ICML}

\vskip 0.3in
]

\printAffiliationsAndNotice{\icmlEqualContribution}

\newcommand{\fix}{\marginpar{FIX}}
\newcommand{\new}{\marginpar{NEW}}

\begin{abstract}
Large Vision-Language Models (LVLMs) typically process visual inputs as a prefix to the language decoder. As the model autoregressively generates text, this initial visual information inevitably undergoes ``dilution'', leading the model to over-rely on language priors and hallucinate objects. Existing interventions attempt to correct this by contrasting logits or iteratively refining outputs, but they incur prohibitive latency costs. We propose Residual-Update Directed DEcoding Regulation \texttt{(RUDDER)}, a framework that counters visual dilution by creating a persistent visual anchor. We extract a robust evidence direction (\texttt{CARD}) directly from the model's prefill residual updates, and inject it into the decoding process. This injection is modulated by an adaptive gate, the \texttt{Beta Gate}, which acts as a trust mechanism and ensures the visual reminder is applied only when necessary. Experiments on LLaVA-1.5 (7B/13B), Idefics2, InstructBLIP, and Qwen2.5-VL demonstrate that RUDDER consistently mitigates hallucination (with greedy decoding, RUDDER reduces CHAIR$_S$ by an average of 24.4\% and CHAIR$_i$ by 23.6\% relative) and scales effectively across architectures, all while maintaining \>96.0\% throughput. The code is available at \href{https://github.com/Akko000/RUDDER-Residual-Update-Directed-DEcoding-Regulation-}{RUDDER}.
\end{abstract}

\section{Introduction}
\label{sec:intro}

\begin{figure*}[t]
  \centering
  \includegraphics[width=\linewidth,keepaspectratio]{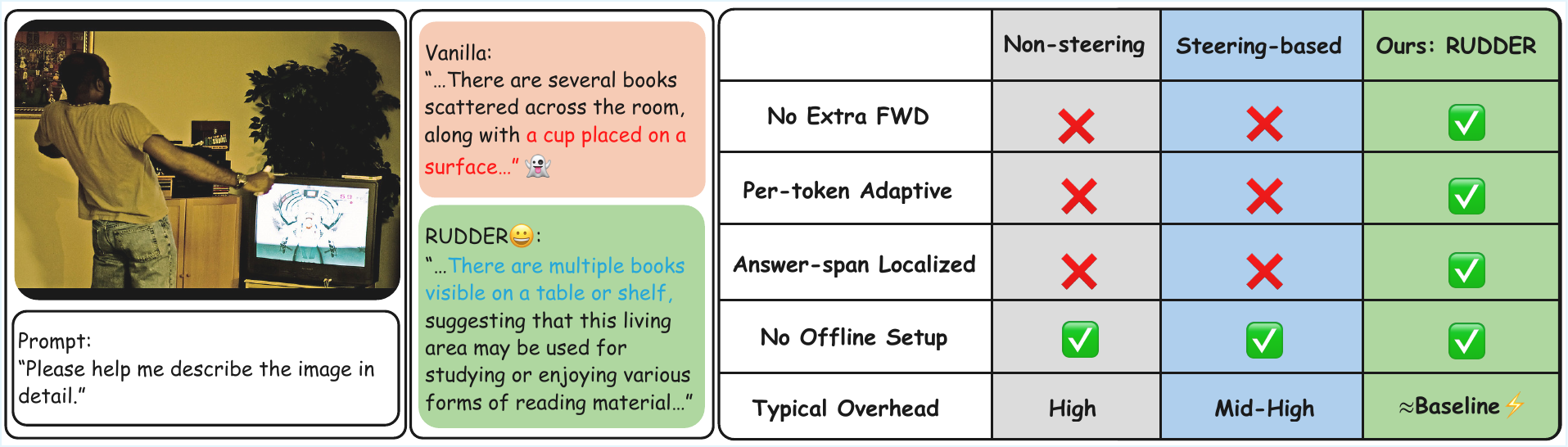}
  \caption{\textbf{(Left)} An example where the vanilla LLaVA-1.5--7B~\citep{liu_improved_2024} hallucinates objects. Erroneous text is marked in \textcolor{red}{red}, while RUDDER's corrected, factual output is in \textcolor{cyan}{blue}. \textbf{(Right)} A comparison showing that unlike existing non-steering and steering-based methods, RUDDER provides adaptive, low-overhead control without requiring extra forward passes.}
  \label{fig:motivation}
  \vspace{-8pt}
\end{figure*}

Large Vision-Language Models (LVLMs) usually encode visual inputs as a prefix to the language decoder. Since the model generates text autoregressively, this initial visual information inevitably undergoes ``dilution''~\citep{liu_paying_2024}, resulting in the model over-relying on internal language priors~\citep{leng_mitigating_2023}. This phenomenon is a primary driver of object hallucination. As illustrated in Figure~\ref{fig:motivation} (Left), the model drifts from the visual context, hallucinating a ``cup'' based on the linguistic association with ``surface,'' despite no such object being present.

To mitigate this, existing Inference-Time Intervention (ITI) methods usually employ Contrastive Decoding~\citep{leng_mitigating_2023} or Iterative Refinement~\citep{degf}. However, as summarized in Figure~\ref{fig:motivation}(Right), these approaches face a severe bottleneck: they typically require multiple forward passes (extra FWD) or external classifiers, incurring prohibitive latency costs. This undermines the efficiency of the autoregressive loop, making real-world deployment difficult.

To resolve this efficacy-efficiency trade-off, we propose a direct solution: \textit{Instead of correcting the model after it drifts, we structurally prevent the dilution of visual information.}
Our key insight is that the visual signal is maximally preserved in the residual stream during the mandatory prefill stage. By capturing this signal before it fades, we can create a lightweight ``visual anchor'' without any additional forward pass.

We introduce Residual-Update Directed DEcoding Regulation \texttt{(RUDDER)}. RUDDER extracts the visual evidence as a Contextual Activation Residual Direction \texttt{(CARD)} vector from the prefill stage. During decoding, RUDDER continuously ``reminds'' the LLM of this evidence by injecting the CARD vector into the hidden states.
To ensure this injection is context-aware (``Per-token Adaptive'' feature in Figure~\ref{fig:motivation}), we employ \texttt{Beta Gate}, an adaptive gate based on the Beta Distribution. This gate functions as a trust mechanism: it dynamically reinforces the injection when the generation aligns with the visual anchor, and suppresses it when the model is processing non-visual (syntactic) tokens.

Our contributions are:
\begin{enumerate}
    \item We propose CARD, a method to extract a persistent visual anchor from prefill residual updates. CARD keeps the model grounded in the visual context throughout the decoding process.
    \item We design the Beta Gate, a training-free adaptive gating mechanism based on the Beta Distribution. It ensures that the visual reminder is applied precisely, balancing grounding with fluency.
    \item We demonstrate that RUDDER is general and works with different base models. It scales to larger models, such as LLaVA-1.5-13B, and generalizes to modern architectures with distinct fusion strategies, such as Qwen2.5-VL and InstructBLIP.
    \item Our method introduces little overhead to the generation process. RUDDER operates within a single forward pass with negligible latency ($<4\%$), significantly outperforming multi-pass methods.
\end{enumerate}

\paragraph{Conflict of Interest Disclosure.}
The authors declare no financial conflicts of interest related to this work.

%%% ME
\section{Related Work}
\label{sec:related_work}
Our research is situated at the intersection of inference-time intervention (ITI) and probabilistic gating.

\textbf{Inference-time intervention.} ITI aims to guide a model's generative behavior without modifying its weights. We group existing methods based on where they act on the computation path.

Non-steering methods operate at the output logits or adjust the decoding strategy. Many of these methods recalibrate final logits to improve visual grounding, but often at the cost of significant latency due to extra forward passes. For instance, VCD~\citep{leng_mitigating_2023} uses perturbed images to create a negative context, PAI~\citep{liu_paying_2024} subtracts unconditional (text-only) logits, and MARINE~\citep{zhao2025mitigating} employs a classifier-free guidance style. 
Recent advancements include HALC~\citep{halc}, which introduces adaptive focal-contrast decoding, and AGLA~\citep{agla}, which mitigates hallucinations by assembling global and local attention features.
Additionally, DeGF~\citep{degf} proposes self-correcting decoding via generative feedback, though such iterative refinement typically incurs higher computational overhead.
Similarly, DoLa~\citep{chuang2023dola} contrasts deep vs. shallow logits to suppress generic text.
More efficient alternatives, such as constrained decoding~\citep{hokamp2017lexicallyconstraineddecodingsequence} or post hoc editing~\citep{manakul2023selfcheckgptzeroresourceblackboxhallucination}, are typically less adaptive.

Steering methods directly modify the hidden representations to guide the generation trajectory. Most of these methods also incur high computational costs on-the-fly. For example, ASD~\citep{su_activation_2025} steers away from a predefined hallucination direction, and VISTA~\citep{li_hidden_2025} injects a signal vector computed from activation differences. VTI~\citep{liu2025reducing} attempts to mitigate this cost by shifting the computational burden to an offline precomputation step.

\textbf{Bayesian and Probabilistic Gating.}
Our work is also inspired by Bayesian and probabilistic gating for uncertainty modeling. This includes concepts from Evidential Deep Learning~\citep{sensoy_evidential_2018}, which frames outputs as parameters of a Dirichlet distribution for uncertainty quantification.
Other relevant work explores stochastic gates. For instance, \cite{yamada2020feature} use stochastic gates based on a relaxation of the Bernoulli distribution for feature selection. More directly related to our method, Beta-LSTM~\citep{song_bivariate_2019} replaces standard sigmoid gates with ones derived from a Beta distribution, validating the use of Bayesian principles in gating mechanisms.

\section{Our Method}
\label{sec:methods}
In autoregressive LVLMs, visual information tends to fade as the generation sequence lengthens, a phenomenon often referred to as visual fading or attention decay~\citep{liu_paying_2024}. However, during the initial prefill stage, the fusion of visual and textual information is at its peak. The residual update from the self-attention sublayer, therefore, encodes the net effect of the visual context on the representation of each text token. We hypothesize that by aggregating these updates over the image tokens and text prompt tokens in the prefill span, we can obtain a robust, per-sample vector that captures the direction of visual evidence for the specific input.
Our empirical analysis supports this: the extracted CARD vector creates a systematic, image-conditioned rotation away from a text-only (language prior) direction, and this rotation aligns coherently with the downstream steering mechanism. This confirms the aggregated updates provide a meaningful directional signal rather than random noise (a detailed visualization and quantification is in Appendix~\ref{app:vis}).

To mitigate object hallucination without the high computational costs of existing steering methods, we present Residual-Update Directed DEcoding Regulation \texttt{(RUDDER)}. RUDDER is a low-overhead framework that adaptively steers LVLMs toward visually-grounded generation by injecting a dynamically derived visual evidence vector. Crucially, it delivers context-specific steering within a single forward pass and introduces negligible latency. RUDDER is training-free; its few deployment hyperparameters are selected once on a small held-out validation set, rather than learned from task-specific supervision.

\subsection{Preliminaries}
The decoder in a Transformer-based LVLM operates on a residual stream, where each sublayer's output (e.g., self-attention of the decoder layer) is added back to its input. This output, termed the residual updates $\Delta^l$, represents the new information contributed at layer $l$. We leverage these updates during the two-stage auto-regressive generation process: \textbf{1. Prefill Stage:} The model processes the prefill span, comprising both image tokens and text prompt tokens, in a single parallel forward pass to populate a Key-Value cache. During this mandatory step, we extract the CARD vector by aggregating the self-attention residual updates across all tokens in the prefill span. \textbf{2. Decoding Stage}: The model generates the output sequentially, one token at a time. It's during this phase that we employ Beta Gate for adaptive steering. 
% are added back to the input, forming a residual stream. Inference begins with a parallel \textbf{prefill} phase, where the entire input prompt is processed to populate a Key-Value cache, followed by a token-by-token decoding phase that utilizes this cache.

\paragraph{Scope of the intervention.}
RUDDER acts in the hidden residual stream rather than at the final logit distribution. Its objective is therefore not to provide a formal guarantee on the token-level likelihood, but to preserve and reassert an input-specific visual evidence direction during decoding. This design matches the failure mode we target: hallucination often arises when the generation trajectory drifts from the visual prefix toward language priors, while the original visual evidence is still present in the model's internal computation during prefill.

\subsection{CARD Vector: A Zero-Cost Per-Sample Evidence Direction}
\label{card}
\textbf{Motivation.}
LVLMs fuse visual and textual information through self-attention. The residual update from the self-attention sublayer, therefore, encodes the net effect of the visual context on the representation of each text token. We hypothesize that by aggregating these updates over the image tokens and text prompt tokens in the prefill span, we can obtain a robust, per-sample vector that captures the direction of visual evidence for the specific input~\citep{liu_improved_2024}.
Our empirical analysis supports this: the extracted CARD vector creates a systematic, image-conditioned rotation away from a text-only (language prior) direction, and this rotation aligns coherently with the downstream steering mechanism. This confirms the aggregated updates provide a meaningful directional signal rather than random noise (a detailed visualization and quantification is in Appendix~\ref{app:vis}).

To identify the optimal layer for extracting the CARD vector, we analyze internal dynamics of LLaVA-1.5--7B~\citep{liu_improved_2024}. We find that intervening in the late decoder layers has the greatest potential to influence the model's final output. Full analysis is provided in Appendix~\ref{sec:llava_dynamics}, Figure~\ref{fig:delta_norm}, \ref{fig:ratio}.
\textbf{Extraction.}
In a single standard prefill pass with the image and text prompt, we place a lightweight, read-only hook at the \emph{target} decoder layer $l$ and cache the self-attention output for each token $i$ in the prefill span $\mathcal{T}_{\text{pre}}$, denoted $\mathbf{A}_i^l$. In a pre-norm decoder, the residual update is simply the attention output,
\begin{equation}
  \Delta_i^l \;=\; \mathbf{A}_i^l,
  \label{eq:delta_t}
\end{equation}
We then pool these updates and apply $L_2$ normalization to obtain a per-sample direction:
\begin{equation}
\begin{gathered}
  \mathbf{v}_{\text{CARD}}
  = \frac{\operatorname{Pool}\big(\{\Delta_i^l\}_{i\in\mathcal{T}_{\text{pre}}}\big)}
         {\big\|\operatorname{Pool}\big(\{\Delta_i^l\}_{i\in\mathcal{T}_{\text{pre}}}\big)\big\|_2},
  \\
  \text{where } \operatorname{Pool}(\cdot)\ \text{can be mean or } \|\Delta_i^l\|\text{-weighted mean}.
\end{gathered}
\label{eq:vcard_pool}
\end{equation}
\textbf{Intuition.} Although pooling aggregates strictly over the prefill span, it naturally prioritizes semantic content. Since self-attention updates are magnitude-heavy on informative tokens and lighter on functional syntax, the weighted pooling effectively filters out noise, distilling a robust, input-specific evidence direction that summarizes the net visual influence. This entire process occurs within the single prefill pass and introduces negligible overhead, as no additional forward pass or calibration is required.

\subsection{Beta Gate: Adaptive Gating Based on the Beta Distribution}
\label{sec:beta_gate}

\paragraph{Motivation: A Trust Mechanism.}
Standard steering often applies a fixed vector, which can be harmful when the visual context is irrelevant (e.g., on syntactic tokens). To address this, we introduce an adaptive gate derived from a probabilistic perspective.
Let $\mathbf{h}_{l,t}$ be the hidden state for generating token $t$ at the target layer $l$. We measure its alignment with the visual evidence via cosine similarity: $s_t=\cos(\mathbf{h}_{l,t},\mathbf{v}_{\text{CARD}})$.

We use the function of a Beta-Bernoulli conjugate posterior to design the gate. We view $s_t$ as providing pseudo-counts for a latent ``visual groundedness'' probability.
Crucially, this acts as a trust mechanism rather than an error detector:
\begin{itemize}
    \item \textbf{High alignment ($s_t \uparrow$):} Indicates the generation is consistent with visual evidence (high trust). The gate increases to reinforce this valid trajectory.
    \item \textbf{Low/Negative alignment ($s_t \downarrow$):} Indicates uncertainty or non-visual context. The gate suppresses intervention to prevent over-steering and preserve fluency.
\end{itemize}
This monotone design is intentional. A high value of $s_t$ means the current hidden state is already compatible with the extracted visual evidence, so reinforcing $\mathbf{v}_{\text{CARD}}$ is unlikely to disrupt generation. Conversely, a low or negative value of $s_t$ suggests either non-visual function-token generation or an unstable trajectory, where aggressive steering can hurt fluency or object recall. The gate therefore controls \emph{trust in the current visual alignment}, rather than measuring deviation magnitude.

\paragraph{Gate Formulation.}
Using this intuition, we calculate the gate parameters using a smooth mapping from $s_t$:
\begin{equation}
\begin{gathered}
  \alpha_t = \operatorname{softplus}(k\,s_t+c),
  \\
  \beta_t = \operatorname{softplus}(-k\,s_t+c),
  \\
  g_t = \frac{\alpha_t}{\alpha_t+\beta_t}.
\end{gathered}
\label{eq:softplus}
\end{equation}
Here, $k$ is a sensitivity hyperparameter controlling the steepness of the response, while $c$ is a concentration parameter typically fixed to a robust default (e.g., $c=1$). We optimize these few parameters on a small held-out validation set (details in Sec.~\ref{sec:exp_setup}).
To ensure stability, we clamp the gate's output to a predefined range, $g_t\in[g_{\min},g_{\max}]$. This prevents the gate from completely shutting off ($g_t{\to}0$) or saturating ($g_t{\to}1$) too readily, making the intervention robust to noise.

For generating each token $t$ in the answer, the final steering update $\mathbf{v}_t^{\text{steer}}$ combines the adaptive gate with a global cap $\alpha_{\max}$:
\begin{equation}
   \mathbf{v}_t^{\text{steer}}
   \;=\;
   \underbrace{\big(\alpha_{\max}\, g_t\big)}_{\text{adaptive strength}}\, \mathbf{v}_{\text{CARD}},
   \label{eq:vtsteer}
\end{equation}
This vector is injected into the residual stream immediately after the Self-Attention($\operatorname{SA}$) operation. The updated hidden state, $\mathbf{h}_{l,t}^{\text{new}}$, is thus computed as:
\begin{equation}
    \mathbf{h}_{l,t}^{\text{new}} \;=\; \big(\mathbf{h}_{l,t}+\operatorname{SA}(\mathbf{h}_{l,t})\big)\;+\;\mathbf{v}_t^{\text{steer}}.
    \label{eq:injection}
\end{equation}

The term $\alpha_{\max}g_t$ represents the adaptive strength of the intervention, ensuring a strong corrective signal is applied only when needed; the injection is restricted to the answer span.

\subsection{Integration and Workflow}
\begin{figure*}[t]
  \centering
  \includegraphics[width=\linewidth,keepaspectratio]{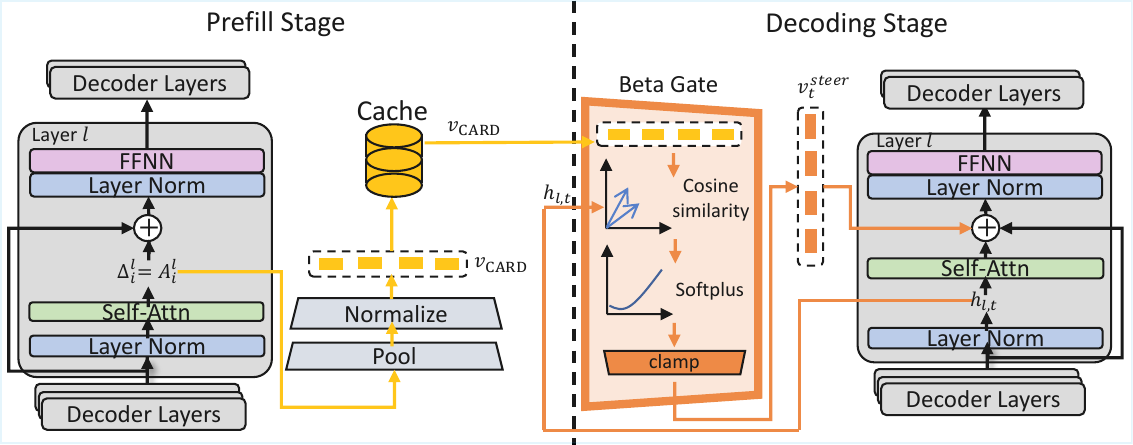}
  \caption{\textbf{The overall workflow of RUDDER.} \textbf{(1) Prefill Stage} (\textcolor{yellow}{Yellow}): We extract the CARD vector $\mathbf{v}_{\mathrm{CARD}}$ by aggregating residual updates from the target layer. \textbf{(2) Decoding Stage} (\textcolor{orange}{Orange}): At each step $t$, the Adaptive Beta Gate computes a steering vector $\mathbf{v}^{\text{steer}}_t$ to modulate the residual stream, guiding the model toward visually grounded outputs.}
%   \caption{\textbf{The overall workflow of RUDDER.} Our method operates in two stages. \textbf{(1) Prefill Stage} (\textcolor{yellow}{Yellow} Arrows): We extract CARD vector $\mathbf{v}_{\mathrm{CARD}}$ by first collecting attention-induced residual updates
% $\Delta_i^{l}$ from a target layer $l$ for each token $i$ in the prefill span. These updates are then aggregated using pooling and normalization. The final CARD vectors are cached for each (image, prompt) pair.
%  (2) \textbf{Decoding Stage} (\textcolor{orange}{Orange} Arrows): When generating each answer token $t$, the adaptive Beta Gate computes a steering vector
% $\mathbf{v}^{\text{steer}}_t$, which is then injected into the residual stream to guide the LVLM towards a more visually-grounded output.}
    
  \label{fig:RUDDER_method}
\end{figure*}
Our complete method, \texttt{Residual-Update Directed DEcoding Regulation (RUDDER)}, integrates the CARD vector and the adaptive Beta Gate to mitigate hallucination by steering LVLMs toward visually grounded outputs. We refer to this primary, adaptive configuration as \texttt{RUDDER-Beta}. To isolate the impact of adaptivity, we also define a static variant, \texttt{RUDDER-Add}, which injects the CARD vector with a constant strength without the gating mechanism.

As detailed in Algorithm~\ref{alg:card}, RUDDER can be seamlessly integrated into the standard auto-regressive decoding loop. By operating within a single inference pass, it mitigates hallucination with negligible computational overhead, resolving the common trade-off between efficacy and efficiency. The overall workflow of this approach is illustrated in Figure~\ref{fig:RUDDER_method}.
% Combining the CARD vector and the Bayesian-inspired gating mechanism is RUDDER(Residual-Update Directed DEcoding Regulation)~\ref{fig:RUDDER_method}. 
% This method utilizes the high leverage found in decoder layers and preserves the stability when directions misalign. As detailed in Algorithm~\ref{alg:card}, RUDDER seamlessly integrates into the standard auto-regressive decoding loop. The CARD vector is extracted once during prefill, while the Beta Gate adaptively applies steering at each subsequent token generation step. RUDDER operates within a \textbf{single-pass inference}, offering a low-overhead solution that resolves the trade-off between high performance and practical efficiency.
% The CARD vector is extracted once during prefill, while the Beta Gate adaptively applies steering at each subsequent token generation step. This approach is effective because it utilizes the high leverage found in mid-to-late decoder layers while preserving stability when internal representations misalign. 
% Ultimately, RUDDER operates within a single-pass inference, offering a low-overhead solution that resolves the trade-off between high performance and practical efficiency. Empirically, the geometry of $\mathbf{v}_t^{\text{steer}}$ and $\mathbf{v}_{\text{CARD}}$ exhibits clear, consistent structure (t\mbox{-}SNE in App.~\ref{app:vis}), supporting the interpretability of our steering signal.
% We achieve hallucination mitigation with a single-pass inference. See Algorithm~\ref{alg:card} for step-by-step pseudo-code.

\section{Experiments}
\label{sec:experiments}

In this section, we validate RUDDER, demonstrating its ability to mitigate hallucination effectively with negligible computational overhead. We conduct a series of experiments across diverse LVLM architectures and benchmarks to assess performance, general capabilities, efficiency, and hyperparameter sensitivity.

\subsection{Experimental Setup}
\label{sec:exp_setup}

\textbf{Model Architectures.} We evaluate RUDDER on three representative LVLMs with distinct visual-textual alignment mechanisms: LLaVA-1.5--7B~\citep{liu_improved_2024} and Idefics2--8B~\citep{laurenccon2024matters} (linear projection), and InstructBLIP~\citep{dai_instructblip_2023} (Q-former). We also include LLaVA-1.5-13B~\citep{liu_improved_2024} and Qwen2.5-VL-7B~\citep{bai2025qwen25vltechnicalreport} in our scalability analysis (Sec.~\ref{sec:scalability}).

\textbf{Baselines.} We compare RUDDER against state-of-the-art inference-time intervention methods. These include logit-based strategies like DoLa~\citep{chuang2023dola}, VCD~\citep{leng_mitigating_2023}, PAI~\citep{liu_paying_2024}, and the recent HALC~\citep{halc}, AGLA~\citep{agla}, and DeGF~\citep{degf}; as well as steering-based interventions like VISTA~\citep{li_hidden_2025}. All baseline results were reproduced under identical evaluation settings.

\textbf{Implementation Details.} We configure RUDDER using a lightweight calibration on 100 held-out MSCOCO images, disjoint from the evaluation subset. This calibration is used only to select deployment hyperparameters and is not a training procedure. The process involves two main steps:
\begin{itemize}
    \item Step 1: Layer Selection ($L$). We perform a linear sweep to maximize CARD vector alignment. We select mid-to-late layers for LLaVA-1.5 ($L=30$) and Idefics2 ($L=28$), and an early layer for InstructBLIP ($L=1$).
    \item Step 2: Hyperparameter Tuning. We grid-search the steering strength $\alpha_{\max}$ and gate sensitivity $k$ to minimize CHAIR scores while maintaining $\ge 95\%$ baseline recall. Other parameters are fixed to robust defaults ($c=1, g \in [0.05, 1]$).
\end{itemize}
Final settings $(\alpha_{\max}, k)$ are $(20, 5.0)$ for LLaVA-1.5, $(8.0, 5.0)$ for Idefics2, and $(6.5, 8.0)$ for InstructBLIP. This one-time setup incurs negligible overhead. We deliberately constrain the search to preserve at least $95\%$ of the vanilla recall, since hallucination reduction is only useful when the model does not lose its general ability to mention correct objects.

\textbf{Evaluation Benchmarks.} We evaluate RUDDER using both specialized hallucination benchmarks and a general capability benchmark.
(1) \textbf{CHAIR}~\citep{rohrbach2019objecthallucinationimagecaptioning}: We evaluate open-ended captioning on MSCOCO~\citep{lin2015microsoftcococommonobjects} validation set, reporting sentence-level ($\mathrm{CHAIR}_{\mathrm{S}}$) and object-level ($\mathrm{CHAIR}_{\mathrm{I}}$) metrics. For the CHAIR benchmark, we evaluate 500 samples using the prompt ``\texttt{Please help me describe the image in detail}'' with a maximum length of 512 tokens.
(2) \textbf{POPE}~\citep{li2023evaluating}: We assess object probing via targeted yes/no questions, reporting Accuracy and F1 scores averaged across random, popular, and adversarial splits. For the POPE benchmark, we use the prompt ``\texttt{Is there a <object> in the image?}'' and let the models to answer with ``\texttt{yes}'' or ``\texttt{no}''.
(3) \textbf{MME}~\citep{fu_mme_2024}: We use MME to confirm that our method maintains general multimodal capabilities (e.g., reasoning, counting).
Following established protocols, we set the temperature to 1.0 and test greedy, beam search (beam=5), and nucleus sampling ($p=0.9$).

\subsection{Results on Hallucination Benchmarks}
\label{sec:exp_main_obj}
\subsubsection{CHAIR: Open-Ended Captioning}

On the CHAIR benchmark, which evaluates hallucination in open-ended captioning, RUDDER demonstrates a strong ability to reduce factual errors while preserving caption quality.

A key challenge in hallucination mitigation is the trade-off with recall: aggressive steering can artificially lower hallucination scores by producing overly simplistic captions. To ensure a fair and practical evaluation, we constrain our analysis to configurations that maintain at least 95\% of the vanilla model's recall (i.e., $\mathrm{Recall}_{\text{\{evaluated methods\}}} \ge 0.95 \times \mathrm{Recall}_{\text{\{vanilla model\}}}$).

Under this constraint, RUDDER-Beta consistently outperforms the vanilla baseline across all tested LVLMs and decoding strategies, as shown in Table~\ref{tab:chair_main}. It achieves average relative reductions of \(\mathbf{33.2\%}\) in sentence-level (CHAIR$_\text{S}$) and \(\mathbf{28.6\%}\) in object-level (CHAIR$_\text{I}$) hallucination across these three models. 

Compared to strong baselines like VCD and DoLa, our method is consistently superior on both metrics. Furthermore, RUDDER-Beta performs on par with the state-of-the-art VISTA and, on average, yields a greater reduction in object-level hallucinations (CHAIR$_\text{I}$).  

RUDDER-Beta's ability to reduce CHAIR$_\text{I}$ more effectively than CHAIR$_\text{S}$ highlights its precision. We attribute this to the token-wise gating mechanism, which selectively amplifies corrections on visually incongruent or content-noun tokens while leaving already grounded tokens largely unperturbed. This allows RUDDER to preferentially suppress object-level hallucinations without degrading overall caption quality and recall.

%\textbf{POPE.} 
\subsubsection{POPE: Visual Question Answering}
Moving from open-ended captioning to a more constrained task, we next evaluate RUDDER on the POPE benchmark for object probing. This benchmark tests the model's factuality through targeted yes/no questions, offering a different perspective on hallucination. In this setting, RUDDER again demonstrates competitive performance.  As shown in Table~\ref{tab:pope}, RUDDER consistently outperforms the vanilla baselines and most competing methods across all tested models. Concretely, RUDDER-Beta improves accuracy by \textbf{\(1.0/0.7/0.5\)} absolute points (pp) and F1 by \textbf{\(1.6/1.3/0.14\)} pp on LLaVA-1.5, Idefics2, and InstructBLIP, respectively.

Notably, RUDDER-Beta achieves the highest F1-score and accuracy on both LLaVA-1.5 and Idefics2, surpassing strong steering-based methods like VISTA. While its performance on InstructBLIP is slightly surpassed by VISTA when employing greedy decoding and nucleus sampling, RUDDER remains highly competitive, highlighting its effectiveness as a versatile solution for reducing object hallucination.

\subsubsection{Analysis of Adaptive vs. Fixed-Strength Steering}
% An internal comparison between our adaptive (-Beta) and fixed (-Add) variants further underscores the value of precision established in the CHAIR benchmark.

A key design choice in RUDDER is the regulation of injection strength. We analyze the trade-off between the adaptive gate (\texttt{RUDDER-Beta}) and the fixed-strength injection (\texttt{RUDDER-Add}). Our experiments show a clear distinction between these variants, guiding the choice based on the task and model architecture.

For complex, open-ended generation (CHAIR), \texttt{RUDDER-Beta} is consistently superior. Its token-wise precision is crucial for suppressing specific hallucinations in long-form text without harming overall recall. In the simpler, binary-choice POPE task, the distinction is more nuanced. While RUDDER-Beta remains the top performer on LLaVA-1.5 and Idefics2, RUDDER-Add is competitive and even surpasses RUDDER-Beta on InstructBLIP.
% RUDDER-Beta's general superiority validates that its adaptive, token-wise gating is crucial for the nuanced demands of both open-ended generation and object probing on models like LLaVA-1.5 and Idefics2. The fixed RUDDER-Add variant, however, shows a competitive edge specifically on InstructBLIP for the simpler POPE task. 
We hypothesize this is partly because InstructBLIP's Q-Former provides a highly-condensed visual representation that responds well to a uniform steering signal in a simple setting. For single-token ``yes/no" answers, the aggressive push from fixed-strength steering can be sufficient and sometimes more beneficial for certain model architectures.
% For POPE's binary decision-making, this already-refined signal benefits more from a simple, uniform boost (Add), making the fine-grained control of Beta less critical in this specific context.

In summary, RUDDER-Beta is recommended for robust and precise control in complex tasks, while the simpler RUDDER-Add is a powerful option for constrained tasks and certain model architectures.

\begin{table*}[ht!]
    \centering
    \caption{\textbf{Hallucination evaluation on the CHAIR benchmark.} We compare RUDDER against state-of-the-art training-free methods with a max generation length of 512 tokens. Note that methods marked with $^{\dagger}$ (e.g., DeGF) require iterative updates, incurring significantly higher latency ($>3\times$) compared to RUDDER's single-pass efficiency. Best results are \textbf{bolded} and second-best are \underline{underlined}.}
    \label{tab:chair_main}
    \resizebox{\textwidth}{!}{
    \begin{tabular}{llcccccc} 
        \toprule
        \multirow{2}{*}{Decoding} & \multirow{2}{*}{Method} & \multicolumn{2}{c}{LLaVA-1.5~\citep{liu_improved_2024}} & \multicolumn{2}{c}{Idefics2~\citep{laurenccon2024matters}} & \multicolumn{2}{c}{InstructBLIP~\cite{dai_instructblip_2023}} \\
        \cmidrule(lr){3-4} \cmidrule(lr){5-6} \cmidrule(lr){7-8}
        & & CHAIR$_\text{S}\downarrow$ & CHAIR$_\text{I}\downarrow$ & CHAIR$_\text{S}\downarrow$ & CHAIR$_\text{I}\downarrow$ & CHAIR$_\text{S}\downarrow$ & CHAIR$_\text{I}\downarrow$ \\
        \midrule
        
        % --- Greedy ---
        \multirow{9}{*}{Greedy}
        & Vanilla & 48.6 & 13.6 & 46.6 & 14.9 & 39.2 & 12.8 \\
        & DoLa~\citep{chuang2023dola} & 47.6 & 13.4 & - & - & - & - \\
        & VCD~\citep{leng_mitigating_2023} & 49.8 & 14.5 & - & - & 46.4 & 15.3 \\
        & HALC~\citep{halc} & 40.3 & 10.3 & - & - & - & - \\
        & AGLA~\citep{agla} & 44.6 & 12.1 & - & - & - & - \\
        & DeGF$^{\dagger}$~\citep{degf} & \textbf{34.2} & \textbf{9.2} & - & - & - & - \\
        & VISTA~\citep{li_hidden_2025} & \underline{38.6} & 11.4 & 33.5 & \underline{11.6} & \underline{27.7} & \underline{9.7} \\
        &\cellcolor{lightgray}\textbf{RUDDER-Beta} (Ours) &\cellcolor{lightgray} 39.5 &\cellcolor{lightgray} \underline{10.5} &\cellcolor{lightgray}\textbf{28.4} &\cellcolor{lightgray}\textbf{10.9} &\cellcolor{lightgray} \textbf{27.1} &\cellcolor{lightgray} \textbf{8.5} \\
        &\cellcolor{lightgray}\textbf{RUDDER-Add} (Ours) &\cellcolor{lightgray} 42.1 &\cellcolor{lightgray} 11.8 &\cellcolor{lightgray}\underline{30.1} &\cellcolor{lightgray}11.8 &\cellcolor{lightgray}28.3 &\cellcolor{lightgray}10.4 \\
        \midrule
        
        % --- Beam Search ---
        \multirow{5}{*}{Beam Search} 
        & Vanilla & 52.8 & 15.6 & 48.6 & 14.5 & 38.2 & 12.7 \\
        & VCD~\citep{leng_mitigating_2023} & 52.4 & 15.5 & - & - & 47.4 & 16.3 \\
        & VISTA~\citep{li_hidden_2025} & \underline{33.9} & \underline{10.5} & 32.2 & 11.8 & \underline{27.1} & \underline{9.6} \\
        & \cellcolor{lightgray}\textbf{RUDDER-Beta} (Ours) &\cellcolor{lightgray}\textbf{33.1} &\cellcolor{lightgray}\textbf{9.3} &\cellcolor{lightgray}\textbf{29.2} &\cellcolor{lightgray}\textbf{10.1} &\cellcolor{lightgray}\textbf{26.2} &\cellcolor{lightgray}\textbf{9.5} \\
        & \cellcolor{lightgray}\textbf{RUDDER-Add} (Ours) &\cellcolor{lightgray}35.2 &\cellcolor{lightgray}10.6 &\cellcolor{lightgray}\underline{31.4} &\cellcolor{lightgray}\underline{10.9} &\cellcolor{lightgray}27.4 &\cellcolor{lightgray}11.1 \\
        \midrule
        
        % --- Nucleus Sampling ---
        \multirow{6}{*}{\shortstack{Nucleus\\Sampling}} 
        & Vanilla & 55.6 & 16.0 & 53.8 & 16.7 & 46.0 & 16.2 \\
        & DoLa~\citep{chuang2023dola} & 49.3 & 14.8 & - & - & - & - \\
        & VCD~\citep{leng_mitigating_2023} & 57.5 & 17.2 & - & - & 53.3 & 19.8 \\
        & VISTA~\citep{li_hidden_2025} & \textbf{39.2} & \underline{11.9} & \underline{35.5} & \underline{11.8} & \underline{29.0} & \textbf{11.3} \\
        & \cellcolor{lightgray}\textbf{RUDDER-Beta} (Ours) &\cellcolor{lightgray}\underline{39.9} &\cellcolor{lightgray}\textbf{11.0} &\cellcolor{lightgray}\textbf{34.1} &\cellcolor{lightgray}\textbf{11.3} &\cellcolor{lightgray}\textbf{28.9} &\cellcolor{lightgray}\underline{13.7} \\
        & \cellcolor{lightgray}\textbf{RUDDER-Add} (Ours)  &\cellcolor{lightgray}41.6 &\cellcolor{lightgray}12.1 &\cellcolor{lightgray}36.5 &\cellcolor{lightgray}12.9 &\cellcolor{lightgray}30.1 &\cellcolor{lightgray}14.4 \\
        \bottomrule
    \end{tabular}
    }
%\vspace{-10pt}
\end{table*}
% Table for POPE results (with aligned numbers)
\begin{table*}[ht!]
    \centering
    \caption{\textbf{Performance on the POPE benchmark across three LVLMs.} The reported values are the mean accuracy and F1 score, aggregated over the random, popular, and adversarial object splits. The best scores are \textbf{bolded}, and the second best scores are \underline{underlined}.}
    \label{tab:pope}
    \resizebox{\textwidth}{!}{
    \begin{tabular}{llcccccc} 
        \toprule
        \multirow{2}{*}{Decoding} & \multirow{2}{*}{Method} & \multicolumn{2}{c}{LLaVA-1.5~\citep{liu_improved_2024}} & \multicolumn{2}{c}{Idefics2~\citep{laurenccon2024matters}} & \multicolumn{2}{c}{InstructBLIP~\cite{dai_instructblip_2023}} \\
        \cmidrule(lr){3-4} \cmidrule(lr){5-6} \cmidrule(lr){7-8}
        & & Acc $\uparrow$ & F1 $\uparrow$ & Acc $\uparrow$ & F1 $\uparrow$ & Acc $\uparrow$ & F1 $\uparrow$ \\
        \midrule
        % --- Greedy ---
        \multirow{6}{*}{Greedy} 
        & Vanilla                   & 85.34 & 84.91 & 78.40 & 74.86 & 85.74 & 84.75 \\
        & DoLa~\citep{chuang2023dola} & 85.51 & 84.96 & -     & -     & -     & -     \\
        & VCD~\citep{leng_mitigating_2023} & 85.46 & 84.87 & -     & -     & 85.79 & 84.89 \\
        & PAI~\citep{liu_paying_2024}      & 85.98 & 85.31 & -     & -     & -     & -     \\
        & VISTA~\citep{li_hidden_2025}     & \underline{86.21} & \underline{85.42} & 78.28     & 74.66     & \textbf{86.25} & \textbf{85.06} \\
        & \cellcolor{lightgray}\textbf{RUDDER-Beta} (Ours) &\cellcolor{lightgray} \textbf{86.53} &\cellcolor{lightgray} \textbf{86.03} &\cellcolor{lightgray} \textbf{78.74} &\cellcolor{lightgray} \textbf{76.52} &\cellcolor{lightgray} 86.02 &\cellcolor{lightgray} 84.93 \\
        & \cellcolor{lightgray}\textbf{RUDDER-Add} (Ours)  &\cellcolor{lightgray} 85.92 &\cellcolor{lightgray} 84.98 &\cellcolor{lightgray} \underline{78.43} &\cellcolor{lightgray} \underline{75.91} &\cellcolor{lightgray} \underline{86.05} &\cellcolor{lightgray} \underline{85.05} \\
        \midrule
        % --- Beam Search ---
        \multirow{6}{*}{Beam Search} 
        & Vanilla                   & 85.46 & 84.98 & 78.67 & 77.55 & 84.73 & 84.37 \\
        & VCD~\citep{leng_mitigating_2023} & 85.60 & 85.06 & -     & -     & 84.95 & 84.59 \\
        & PAI~\citep{liu_paying_2024}      & 85.58 & 85.01 & -     & -     & -     & -     \\
        & VISTA~\citep{li_hidden_2025}     & \underline{86.10} & \underline{85.35} & 78.40     & 77.31     & \underline{85.64} & \underline{84.61} \\
        & \cellcolor{lightgray}\textbf{RUDDER-Beta} (Ours) &\cellcolor{lightgray} \textbf{86.51} &\cellcolor{lightgray} \textbf{86.19} &\cellcolor{lightgray} \textbf{79.33} &\cellcolor{lightgray} \textbf{77.96} &\cellcolor{lightgray} 85.54 &\cellcolor{lightgray} 84.40 \\
        & \cellcolor{lightgray}\textbf{RUDDER-Add} (Ours)  &\cellcolor{lightgray} 85.98 &\cellcolor{lightgray} 85.02 &\cellcolor{lightgray} \underline{78.91} &\cellcolor{lightgray} \underline{77.60} &\cellcolor{lightgray} \textbf{85.71} &\cellcolor{lightgray} \textbf{84.75} \\
        \midrule
        % --- Nucleus Sampling ---
        \multirow{6}{*}{\shortstack{Nucleus\\Sampling}} 
        & Vanilla                   & 83.00 & 81.08 & 74.84 & 67.78 & 85.50 & 84.52 \\
        & DoLa~\citep{chuang2023dola} & 82.94 & 81.12 & -     & -     & -     & -     \\
        & VCD~\citep{leng_mitigating_2023} & 82.82 & 81.90 & -     & -     & 85.61 & 84.65 \\
        & PAI~\citep{liu_paying_2024}      & 83.17 & 82.14 & -     & -     & -     & -     \\
        & VISTA~\citep{li_hidden_2025}     & \underline{83.58} & 82.21 & 74.66     & 67.70     & \textbf{86.12} & \textbf{85.26} \\
        & \cellcolor{lightgray}\textbf{RUDDER-Beta} (Ours) &\cellcolor{lightgray} \textbf{84.02} &\cellcolor{lightgray} \textbf{83.57} &\cellcolor{lightgray} \textbf{75.89} &\cellcolor{lightgray} \textbf{69.69} &\cellcolor{lightgray} 85.79 &\cellcolor{lightgray} 84.74 \\
        & \cellcolor{lightgray}\textbf{RUDDER-Add} (Ours)  &\cellcolor{lightgray} 83.20 &\cellcolor{lightgray} \underline{82.38} &\cellcolor{lightgray} \underline{74.95} &\cellcolor{lightgray} \underline{67.84} &\cellcolor{lightgray} \underline{85.95} &\cellcolor{lightgray} \underline{84.95} \\
        \bottomrule
    \end{tabular}
    }
\end{table*}
\subsection{Results on Comprehensive Benchmarks}
\label{sec:exp_main_comp}
% After validating RUDDER on hallucination benchmarks, we still need to confirm that RUDDER does not sacrifice the overall capabilities to reduce hallucination. 
To ensure that hallucination mitigation does not compromise general multimodal capabilities, we evaluate RUDDER on \textbf{MME} benchmark. The results show that RUDDER successfully reduces hallucinations without sacrificing the overall abilities of the tested LVLMs.
% \textbf{MME Evaluation.} We conduct a full-set evaluation across all LVLM architectures and decoding strategies. 
As demonstrated in Table~\ref{tab:mme}, both RUDDER-Beta and RUDDER-Add achieve higher MME scores than the vanilla models for Idefics2 and InstructBLIP. On LLaVA-1.5, RUDDER's scores are slightly lower than the vanilla model, but the difference is still acceptable. 
% which could be the result of high $\alpha_{\text{max}}$. 
% This means RUDDER reduces hallucinations without compromising the overall abilities of LVLMs.

\begin{table}[h]
    \centering
    \caption{\textbf{Overall performance scores on the MME full evaluation set.} Higher scores indicate better general capability across perception, reasoning, and knowledge-based tasks.}
    \label{tab:mme}
    \resizebox{\linewidth}{!}{
    \begin{tabular}{llccc}
        \toprule
        Decoding & Method & LLaVA-1.5 & Idefics2 & InstructBLIP \\
        \midrule
        \multirow{3}{*}{Greedy} 
        & Vanilla & 1745.87 & 1518.84 & 1566.77 \\
        & \textbf{RUDDER-Beta} & 1724.17 & 1540.56 & 1592.07 \\
        & \textbf{RUDDER-Add} & 1715.45 & 1526.03 & 1585.28 \\
        \midrule
        \multirow{3}{*}{Beam Search} 
        & Vanilla & 1760.20 & 1450.59 & 1539.16 \\
        & \textbf{RUDDER-Beta} & 1746.66 & 1484.21 & 1565.77 \\
        & \textbf{RUDDER-Add} & 1738.13 & 1475.80 & 1560.64 \\
        \midrule
        \multirow{3}{*}{\shortstack{Nucleus\\Sampling}} 
        & Vanilla & 1752.65 & 1362.45 & 1538.18 \\
        & \textbf{RUDDER-Beta} & 1721.94 & 1374.77 & 1556.43 \\
        & \textbf{RUDDER-Add} & 1713.74 & 1364.16 & 1546.01 \\
        \bottomrule
    \end{tabular}
    }
\end{table}

\subsection{Efficiency Tests}
\label{sec:ablation}
A critical advantage of RUDDER is its low computational overhead, making it practical for real-world deployment. Unlike many state-of-the-art intervention methods that require extra forward passes and significantly increase latency, RUDDER is designed to operate within a single generative pass. We measure the practical latency and throughput of RUDDER against vanilla models and other methods, with results presented in Table~\ref{tab:throughput}. All experiments are conducted on a single Nvidia A100 GPU with 80 GB VRAM and a batch size fixed at 1. RUDDER-Beta maintains an average throughput of 96.0\% compared to vanilla LVLMs. RUDDER-Add is even more efficient as it bypasses the Beta Gate calculation. In contrast, competing methods that require extra forward passes see a significant drop in efficiency. On average, the throughput of a method like VISTA is only 58.1\% of the vanilla models.
\begin{table}[h!]
    \centering
    \caption{\textbf{Throughput Comparison (tokens/s).} Measurements are conducted using greedy decoding. Higher scores indicate better efficiency.}
    \label{tab:throughput}
    % 保持使用 \resizebox{\linewidth}，但因为内容变窄了，缩放比例会变大，字体就会变大
    \resizebox{\linewidth}{!}{
    \begin{tabular}{lccc}
        \toprule
        Method & LLaVA-1.5 & Idefics2 & InstructBLIP \\ % 移除了表头的引用
        \midrule
        Vanilla & 56.7 & 47.8 & 62.3 \\
        VCD     & 30.1 & -    & -    \\ % 移除了方法列的引用
        PAI     & 29.5 & -    & -    \\
        VISTA   & 36.1 & 31.9 & 28.9 \\
        \cellcolor{lightgray}\textbf{RUDDER-Beta} (Ours) & \cellcolor{lightgray}54.9 & \cellcolor{lightgray}45.8 & \cellcolor{lightgray}59.5 \\
        \cellcolor{lightgray}\textbf{RUDDER-Add} (Ours)  & \cellcolor{lightgray}55.8 & \cellcolor{lightgray}46.5 & \cellcolor{lightgray}60.8 \\
        \bottomrule
    \end{tabular}
    }
    %\vspace{-10pt}
\end{table}

\subsection{Ablation Studies}

% --- 关键修改：把 Figure 3 (单栏图) 的代码提上来，并把参数改成 [t] ---
\begin{figure}[h] % 绝对不要用 [h]，改成 [t] (Top) 或 [tb]
  \centering
  \includegraphics[width=\linewidth,keepaspectratio]{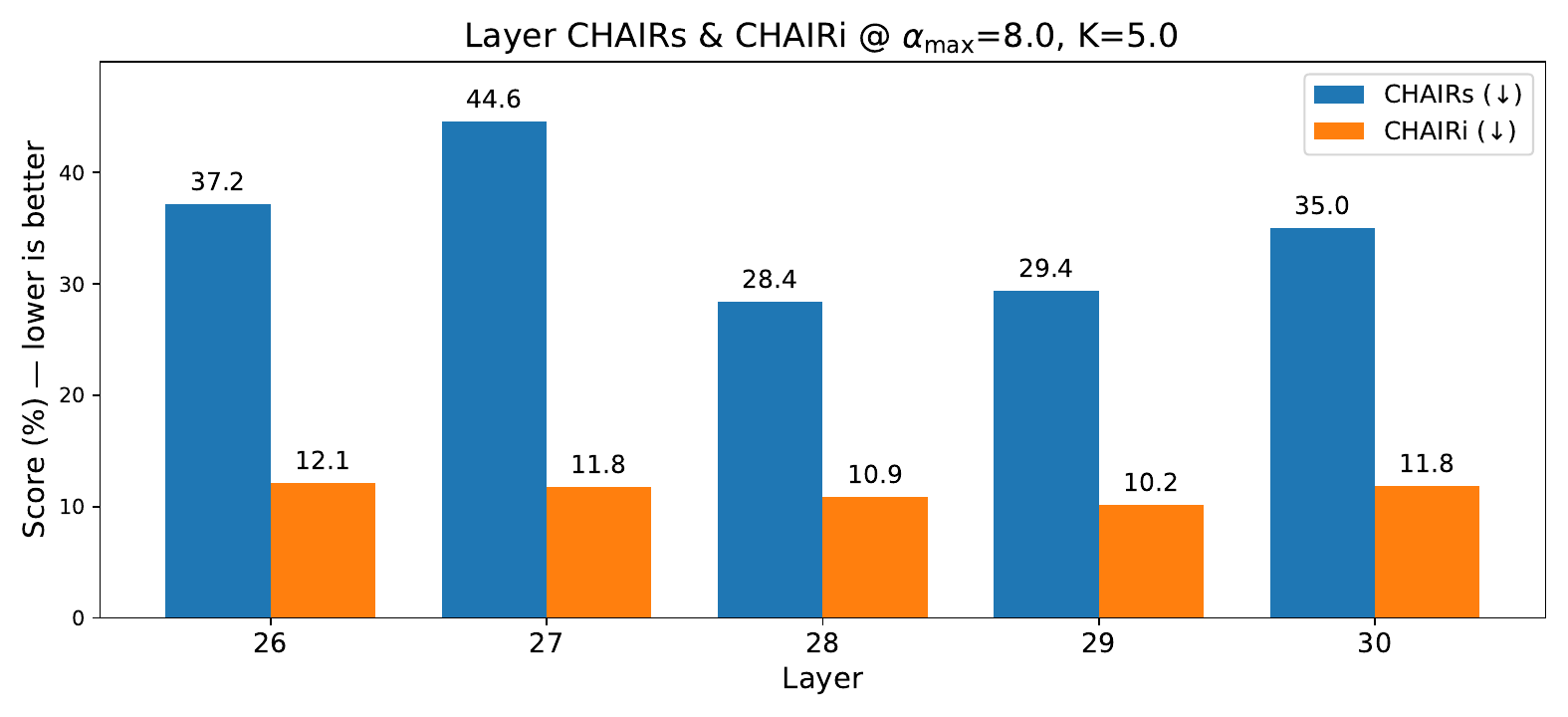}
  %\vspace{-10pt} % 稍微减少图与 caption 的间距
  \caption{\textbf{Layer ablation study on Idefics2.} We fix $(\alpha_{\max}=8.0,\;k=5.0)$ and vary the injection layer. Mid–late layers ($L\!\approx\!28$–$30$) prove most effective, with $L{=}28$ yielding a strong reduction.}
  \label{fig:layer_bar}
  \vspace{-10pt} % 减少图与下方正文的间距
\end{figure}
% -------------------------------------------------------------------

% --- Figure 4 (跨栏大图) 保持 [t] ---
\begin{figure*}[t] 
  \centering
  \begin{subfigure}[t]{0.32\linewidth}
    \centering
    \includegraphics[width=\linewidth,keepaspectratio]{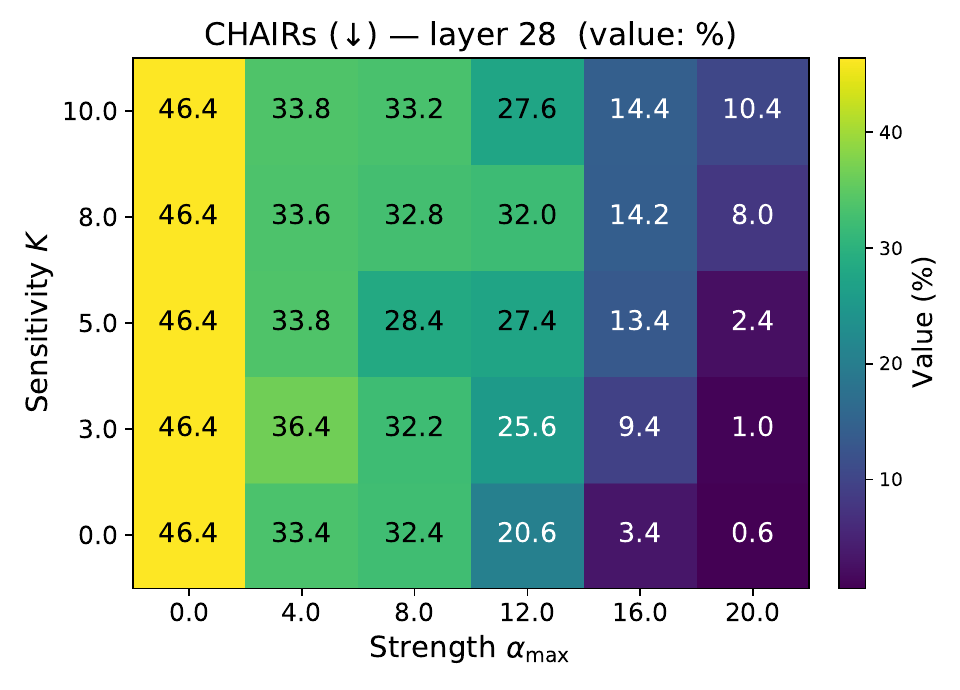}
    \caption{CHAIR$_\text{S}$ heatmap}
    \label{fig:heatmap_chairs}
  \end{subfigure}\hfill
  \begin{subfigure}[t]{0.32\linewidth}
    \centering
    \includegraphics[width=\linewidth,keepaspectratio]{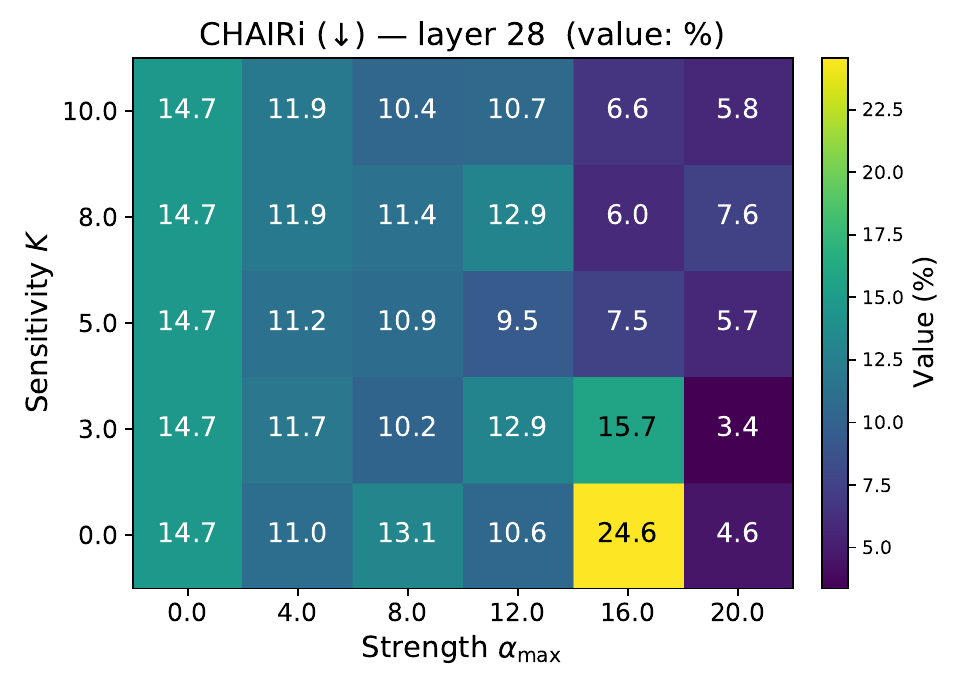}
    \caption{CHAIR$_\text{i}$ heatmap}
    \label{fig:heatmap_chairi}
  \end{subfigure}\hfill
  \begin{subfigure}[t]{0.32\linewidth}
    \centering
    \includegraphics[width=\linewidth,keepaspectratio]{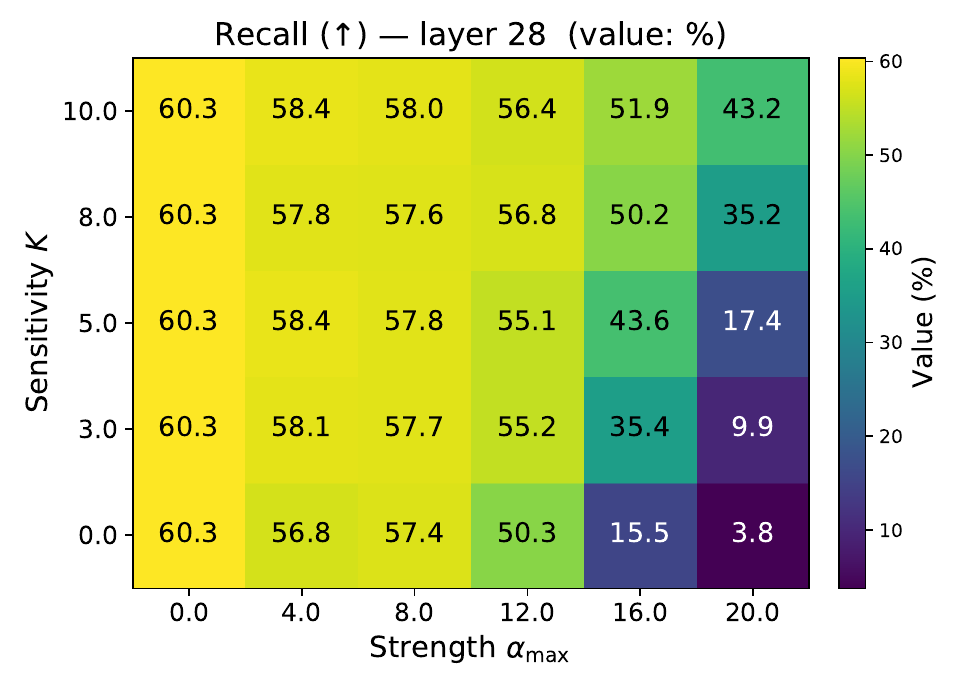}
    \caption{$Recall$ heatmap}
    \label{fig:heatmap_recall}
  \end{subfigure}
  \vspace{-5pt}
  \caption{\textbf{Hyperparameter sensitivity analysis on Idefics2.} We analyze the trade-off between steering strength $\alpha_\text{max}$ and gate sensitivity $k$ at Layer 28. Stronger steering reduces hallucinations (a, b) but may impact recall (c). The optimal balance is found around $\alpha_{\max}=8.0, k=5.0$.}
  \label{fig:ablation_heatmaps}
  %\vspace{-10pt}
\end{figure*}
% ------------------------------------------------

We conduct an ablation study on Idefics2 using the CHAIR benchmark to analyze the key hyperparameters: injection layer $L$, maximum steering strength $\alpha_{\text{max}}$ and the gate sensitivity $k$. 

We first identify the optimal intervention layer, finding Layer 28 is the most effective for the Idefics2 model, as shown in Figure~\ref{fig:layer_bar}. 
Focusing on this layer, we then tune the hyperparameters $\alpha_{\text{max}}$ and $k$. 
The heatmaps in Figures~\ref{fig:heatmap_chairs} through \ref{fig:heatmap_recall} reveal a core trade-off: increasing the steering strength ($\alpha_{\text{max}}$) effectively reduces CHAIR scores but at the cost of lower recall. 
The gate sensitivity $k$, does not exhibit a simple linear trend; instead, it plays a crucial modulating role in this trade-off. 
Ultimately, we find that the best balance for Idefics2 is achieved with $\alpha_\text{max}=8.0$ and $k=5.0$. 
Ablation results for other models are presented in Appendix~\ref{sec:additional_ablation}.

\subsection{Scalability and Generalization Analysis}
\label{sec:scalability}

To verify that RUDDER is not over-fitted to specific model sizes or architectures, we extend our evaluation to LLaVA-1.5-13B (to verify scalability) and the recent Qwen2.5-VL-7B (to verify architectural generalization).

As shown in Table~\ref{tab:generalization}, RUDDER demonstrates robust performance across these diverse settings. 
On the larger LLaVA-1.5-13B, RUDDER scales effectively, surpassing the VISTA baseline in sentence-level hallucination reduction (CHAIR$_S$: 39.9 vs. 40.1) and achieving the highest POPE performance (F1: 85.5).

The results on Qwen2.5-VL highlight the precision of our method. While VISTA achieves a lower CHAIR$_S$, it suffers a regression in POPE performance compared to the vanilla baseline (F1: $87.8 \to 87.4$, Acc: $88.8 \to 88.6$), suggesting that its aggressive steering may compromise the model's fundamental object recognition capabilities. 
In contrast, RUDDER achieves the lowest instance-level hallucination rate (CHAIR$_I$: 7.0) while simultaneously improving POPE performance (F1: 88.1). This confirms that the CARD vector captures a universal and precise visual evidence signal, mitigating hallucinations without the side effect of degrading general recognition accuracy.

\begin{table}[h]
    \centering
    \caption{\textbf{Scalability and Generalization Analysis.} We assess RUDDER on a larger backbone (LLaVA-1.5-13B) and a distinct modern architecture (Qwen2.5-VL-7B) using Greedy decoding.}
    \label{tab:generalization}
    
    % 设置字号为 small，增加列间距
    \small 
    \setlength{\tabcolsep}{5pt} 
    
    \begin{tabular}{llcccc}
        \toprule
        \multirow{2}{*}{\textbf{Model}} & \multirow{2}{*}{\textbf{Method}} & \multicolumn{2}{c}{\textbf{CHAIR}} & \multicolumn{2}{c}{\textbf{POPE}} \\
        \cmidrule(lr){3-4} \cmidrule(lr){5-6}
         & & C$_S$$\downarrow$ & C$_I$$\downarrow$ & Acc$\uparrow$ & F1$\uparrow$ \\
        \midrule
        
        % --- LLaVA-1.5-13B ---
        \multirow{3}{*}{\shortstack[l]{LLaVA-1.5\\(13B)}} 
        & Vanilla & 44.2 & 11.8 & 85.7 & 85.0 \\
        & VISTA & 40.1 & \textbf{10.6} & \underline{86.0} & \underline{85.2} \\
        & \cellcolor{lightgray}\textbf{RUDDER} & \cellcolor{lightgray}\textbf{39.9} & \cellcolor{lightgray}\underline{10.8} & \cellcolor{lightgray}\textbf{86.3} & 
        \cellcolor{lightgray}\textbf{85.5} \\
        \midrule
        
        % --- Qwen2.5-VL-7B ---
        \multirow{3}{*}{\shortstack[l]{Qwen2.5-VL\\(7B)}} 
        & Vanilla & 35.2 & 9.5 & \underline{88.8} & \underline{87.8} \\
        & VISTA & \textbf{25.1} & \underline{7.7} & 88.6 & 87.4 \\
        & \cellcolor{lightgray}\textbf{RUDDER} & \cellcolor{lightgray}\underline{26.9} & \cellcolor{lightgray}\textbf{7.0} & \cellcolor{lightgray}\textbf{89.0} & \cellcolor{lightgray}\textbf{88.1} \\
        \bottomrule
    \end{tabular}
\end{table}
\subsection{Case Study}
Qualitative analysis in Appendix~\ref{sec:case_study} demonstrates RUDDER's effectiveness. The case studies show that RUDDER not only eliminates object hallucinations present in the vanilla model's outputs but also produces more conservative content. By avoiding the vanilla model’s confident yet incorrect assertions, RUDDER enhances the model's overall reliability.

\section{Conclusion and Limitations}
In this work, we introduce \texttt{RUDDER}, a low-overhead inference-time intervention framework that mitigates LVLMs hallucination using two key innovations: the zero-cost \texttt{CARD vector}, which extracts a per-sample visual evidence from the model's own residual updates, and the adaptive \texttt{Beta Gate}, which applies a corrective signal with principled, token-wise strength.

Experiments confirm RUDDER achieves state-of-the-art comparable performance on benchmarks like CHAIR and POPE with negligible computational overhead, resolving the common efficacy-efficiency trade-off. RUDDER presents a practical and effective solution for enhancing the reliability of LVLMs in real-world settings.

RUDDER's primary limitation is its sensitivity to hyperparameters, which need be tuned for each model architecture. Future work could focus on automated hyperparameter optimization to improve its robustness and ease of deployment.

%\clearpage
\section*{Impact Statement}
This paper presents RUDDER, a method designed to mitigate object hallucinations in LVLMs. Hallucinations pose significant risks in real-world applications, such as autonomous systems, medical imaging analysis, and robotics, where factual errors can lead to safety hazards. By improving the grounding of model outputs in visual evidence, our work contributes to the development of more reliable and trustworthy AI systems.

A core contribution of RUDDER is its computational efficiency. RUDDER operates with negligible overhead. This efficiency aligns with the goals of ``Green AI,'' reducing the energy footprint of high-quality model inference and lowering the barrier to deploying aligned LVLMs in resource-constrained environments.

\section*{Acknowledgements}

This work was supported by the Research Council of Finland 
(Flagship programme: Finnish Center for Artificial Intelligence FCAI, 
and grant 358246).

\bibliography{Steering_Vector}
\bibliographystyle{icml2026}

\appendix
\section{Additional Illustration on the Methodology}
\subsection{RUDDER Algorithm}
Here we present the pseudo-code for RUDDER illustrated in Sec.~\ref{sec:methods}.
\begin{algorithm*}[h]
\caption{RUDDER (single-pass, test-time steering; fixed target layer $\ell$)}
\label{alg:card}
\begin{algorithmic}[1]
% Input
\STATE \textbf{Input:} Model $M$; image $x_{\text{img}}$, text $x_{\text{text}}$; fixed layer $\ell$;
          hyperparams $(\alpha_{\max},k,c,g_{\min},g_{\max})$

% Tokenize
\STATE $\mathcal{T}_{\text{pre}} \leftarrow \mathrm{TokenizePrefill}(x_{\text{img}},x_{\text{text}})$ \COMMENT{image + prompt tokens}

% Prefill
\STATE \textbf{Prefill:} run $M$ once (read-only hook at layer $\ell$) to build KV cache and cache $\{\mathbf{A}^{\ell}_i\}_{i\in\mathcal{T}_{\text{pre}}}$

% Delta & CARD
\STATE $\Delta^{\ell}_i \leftarrow \mathbf{A}^{\ell}_i$ \textbf{by} Eq.~\ref{eq:delta_t}
\COMMENT{pre-norm: residual update equals SA output}

\STATE $\mathbf{v}_{\mathrm{CARD}} \leftarrow$ \textbf{by} Eq.~\ref{eq:vcard_pool}
\COMMENT{Pool $\rightarrow$ $L_2$-Normalize over $\{\Delta^{\ell}_i\}_{i\in\mathcal{T}_{\text{pre}}}$}

% Decode Loop
\STATE \textbf{Decode:} for $t=1,2,\dots$ \COMMENT{auto-regressive generation}
    
    \STATE \hskip1em $s_t \leftarrow \cos\!\big(\mathbf{h}_{\ell,t},\,\mathbf{v}_{\mathrm{CARD}}^{\ell}\big)$

    \STATE \hskip1em $(\alpha_t,\beta_t,g_{\ell,t}) \leftarrow$ \textbf{by} Eq.~\ref{eq:softplus}
    
    \STATE \hskip1em $g_{\ell,t} \leftarrow \mathrm{clip}(g_{\ell,t}, g_{\min}, g_{\max})$

    \STATE \hskip1em $\mathbf{h}_{\ell,t}^{new} \leftarrow \big(\mathbf{h}_{\ell,t}+\mathrm{SA}^{(\ell)}(\mathbf{h}_{\ell,t})\big) 
                  + \mathbf{1}[t\in\mathcal{T}_{\text{ans}}]\cdot \mathbf{v}^{\text{steer}}_{t}$
    \COMMENT{post-SA residual add; \textbf{answer span only}}

    \STATE \hskip1em \textbf{emit} next token 
    
\STATE \textbf{end for}
\end{algorithmic}
\end{algorithm*}
\subsection{From Naïve Bayes to the Adaptive Gate}
\label{app:bayes-gate}

\paragraph{Problem setup.}
At each decoding step $t$, we want a scalar gate $g_t\!\in\!(0,1)$ that reflects how much the current token should be nudged toward the visual evidence direction $\mathbf{v}_{\text{CARD}}$.
Let the alignment statistic be $s_t=\cos(\mathbf{h}_t,\mathbf{v}_{\text{CARD}})\in[-1,1]$.

\paragraph{Naïve Bayes view (posterior as a gate).}
Introduce a latent Bernoulli variable $Z_t\!\in\!\{0,1\}$ indicating whether the token is visually grounded ($Z_t=1$) or at risk of drifting ($Z_t=0$).
We use the posterior mean $g_t=\mathbb{E}[Z_t\mid s_t]$ as a \emph{continuous} gate (rather than a hard on/off decision).

\paragraph{Beta--Bernoulli conjugacy with ``soft counts''.}
With a Beta prior $\mathrm{Beta}(\alpha_t,\beta_t)$ on $Z_t$, the posterior mean is
\[
g_t \;=\; \frac{\alpha_t}{\alpha_t + \beta_t}.
\]
We map the alignment $s_t$ to \emph{positive pseudo-counts} via a smooth, monotone transform:
\[
\alpha_t=\operatorname{softplus}(k\,s_t+c),\qquad
\beta_t=\operatorname{softplus}(-k\,s_t+c),
\]
where $k$ controls sensitivity and $c$ controls concentration/bias. The softplus
ensures strictly positive, numerically stable ``counts''.

\paragraph{Properties (useful for calibration).}
The resulting $g_t$ is monotone in $s_t$, symmetric $g(-s)=1-g(s)$, and bounded in $(0,1)$.
Around $s=0$, the slope
\[
\left.\frac{\partial g}{\partial s}\right|_{s=0}
= \frac{k\,\sigma(c)}{2\,\mathrm{softplus}(c)},\quad
\sigma(x)=\frac{1}{1+e^{-x}},
\]
gives a handy knob to set how fast the gate reacts to alignment changes.

\paragraph{Stability: clamping and optional per-token cap.}
For robustness we clamp $g_t\!\leftarrow\!\mathrm{clip}(g_t; g_{\min}, g_{\max})$ to avoid both shutting off ($g_t\!\to\!0$) and saturating ($g_t\!\to\!1$).
Optionally, we enforce a per-token norm cap $\tau$:
\[
\big\|\alpha_{\max} g_t\,\widehat{\mathbf{v}}_{\text{CARD}}\big\|_2 \le \tau,\qquad
\widehat{\mathbf{v}}=\mathbf{v}/\|\mathbf{v}\|_2,
\]
which further prevents rare spikes when hidden-state norms vary.

\paragraph{Final update (matches Algorithm~\ref{alg:card}).}
\begin{equation}
   \mathbf{v}_t^{\text{steer}}
   \;=\;
   \underbrace{\big(\alpha_{\max}\, g_t\big)}_{\text{adaptive strength}}\, \mathbf{v}_{\text{CARD}}
   % \label{eq:vtsteer}
\end{equation}
\begin{equation}
    \mathbf{h}_{l,t}^{\text{new}} \;=\; \big(\mathbf{h}_{l,t}+\mathrm{SA}(\mathbf{h}_{l,t})\big)\;+\;\mathbf{v}_t^{\text{steer}}
\end{equation}
We apply this only on the answer span using the mask $m_t$ as in Algorithm~\ref{alg:card}.

\paragraph{Implementation notes.}
\begin{itemize}
  \item We compute $s_t$ with L2-normalized $\mathbf{h}_t$ and $\mathbf{v}_{\text{CARD}}$ (cosine similarity).
  \item $g_t$ is clamped to $[g_{\min},g_{\max}]$; the global scale $\alpha_{\max}$ controls the maximal push.
  \item $\mathbf{v}_{\text{CARD}}$ is extracted once during the mandatory prefill pass (zero extra forwards).
\end{itemize}

\paragraph{Practical calibration recipe.}
Choose $c$ to set the overall smoothness (typical $c\!\in\![0.5,1.5]$), then increase $k$ until $g_t$ becomes sufficiently responsive on a small dev set.
Finally tune $\alpha_{\max}$ and $[g_{\min},g_{\max}]$ for stability/strength trade-offs.

\paragraph{Complexity.}
The gate requires only light-weight vector ops during decoding and reuses the prefill to compute $\mathbf{v}_{\text{CARD}}$. Hence no extra forward pass compared to vanilla generation.

\subsection{Clarification}
RUDDER maintains the same inference efficiency as the original model, requiring no extra forward passes. The CARD vector is extracted opportunistically during the mandatory prefill pass, and the steering is applied within each step of the subsequent decoding pass.

By no extra forward pass we mean no additional \underline{model.forward} invocations beyond the vanilla prefill and decode; our overhead comes only from cheap per-token vector operations implemented via hooks.

\subsection{Visualization and Quantification of the Geometry of $\mathbf{v}_{\text{CARD}}$ and $\mathbf{v}_t^{\text{steer}}$}
\label{app:vis}

\begin{figure}[t]
  \centering
  \begin{subfigure}[t]{0.49\linewidth}
    \centering
    \includegraphics[width=\linewidth,keepaspectratio]{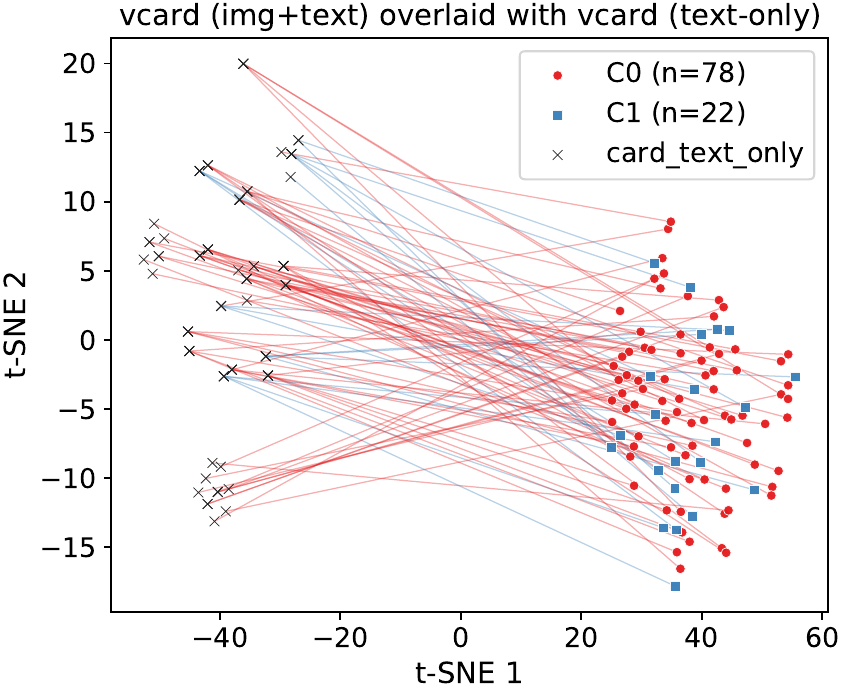}
    \caption{t-SNE overlay of $\mathbf{v}_{\text{CARD}}$ (img+text; colored by $\mathbf{v}_t^{\text{steer}}$ clusters) and text-only $\mathbf{v}_{\text{CARD}}$ (black), with lines linking paired samples.}
    \label{fig:vis_overlay}
  \end{subfigure}\hfill
  \begin{subfigure}[t]{0.49\linewidth}
    \centering
    \includegraphics[width=\linewidth,keepaspectratio]{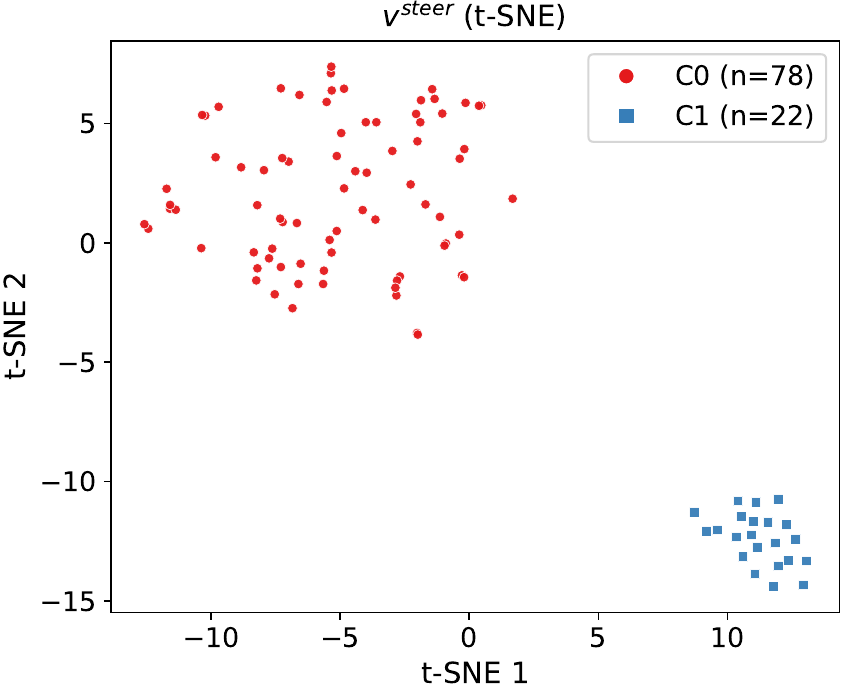}
    \caption{t-SNE of $\mathbf{v}_t^{\text{steer}}$ with KMeans clusters.}
    \label{fig:vis_beta}
  \end{subfigure}
  \caption{Structure in steering space (b) and its sample-wise projection to $\mathbf{v}_{\text{CARD}}$ (a).}
  \label{fig:app_vis_tsne}
\end{figure}

\paragraph{Setup and link to motivation.}
Large VLMs fuse vision and text via self-attention; the residual update of this sublayer thus captures the \emph{net} impact of visual context on token representations.
Motivated by this, we aggregate the prefill-phase residual updates to obtain a per-sample direction $\mathbf{v}_{\text{CARD}}\!\in\!\mathbb{R}^d$ and define its steering counterpart $\mathbf{v}_t^{\text{steer}} \!=\! (\alpha_{\max}\,\overline{\mathrm{gate}})\,\mathbf{v}_{\text{CARD}}$ (Sec.~\ref{sec:methods}).
For each image we export (i) $\mathbf{v}_{\text{CARD}}$ (image+prompt) and its text-only variant, and (ii) $\mathbf{v}_t^{\text{steer}}$.
We reduce vectors by PCA ($k{=}50$) then t\hbox{-}SNE (default perplexity unless noted), and cluster the steering space with KMeans (best silhouette over $K\!\in\!\{2,\dots,10\}$).
Figure~\ref{fig:app_vis_tsne} shows two key views: a \emph{paired} overlay of image+text \emph{vs.} text-only $\mathbf{v}_{\text{CARD}}$ with one-to-one lines, and the clustered t\hbox{-}SNE of $\mathbf{v}_t^{\text{steer}}$. The overlay reveals systematic sample-wise \emph{rotations} from the language prior (text-only) to the image-conditioned direction, and these rotations point toward coherent steering clusters—visual evidence is therefore \emph{directional} rather than noise, directly supporting our motivation.

\begin{figure}[t]
  \centering
  % -------- Row 1: two histograms (fill the row) --------
  \begin{subfigure}[t]{0.495\linewidth}
    \centering
    \includegraphics[width=\linewidth,keepaspectratio]{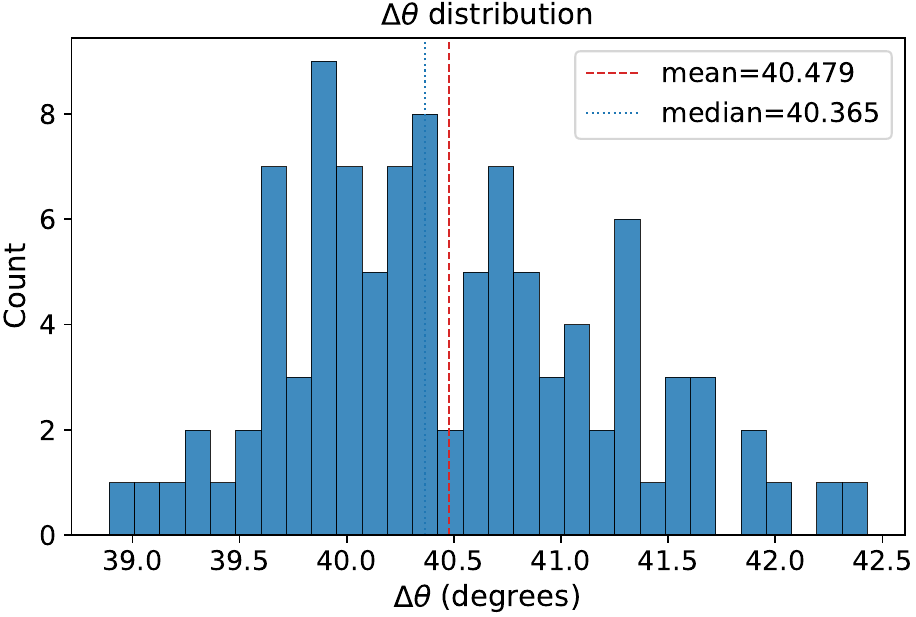}
    \caption{$\Delta\theta$ distribution.}
  \end{subfigure}\hfill
  \begin{subfigure}[t]{0.495\linewidth}
    \centering
    \includegraphics[width=\linewidth,keepaspectratio]{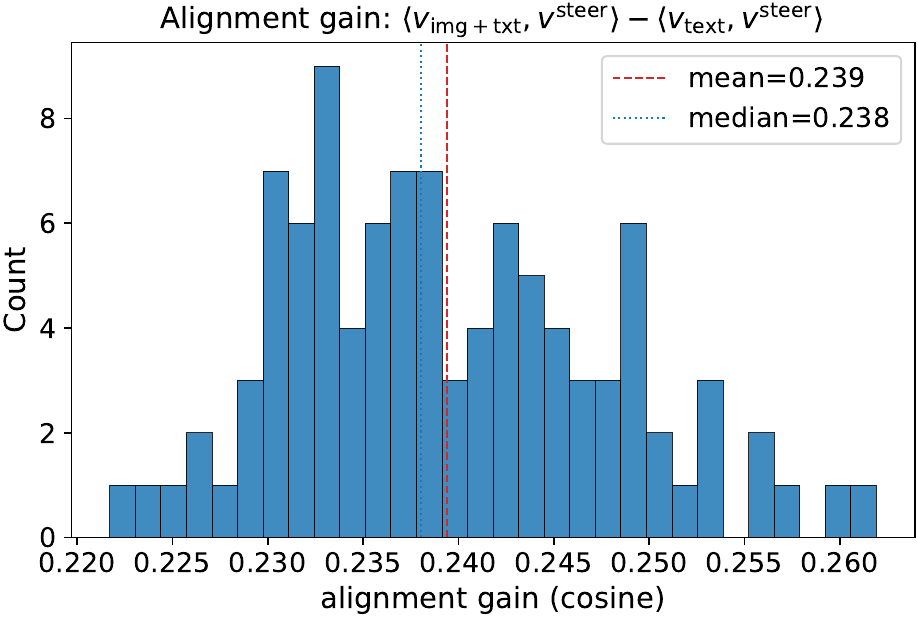}
    \caption{Alignment gain histogram.}
  \end{subfigure}

  \vspace{0.6em}

  % -------- Row 2: three cluster-wise plots (fill the row) --------
  \begin{subfigure}[t]{0.32\linewidth}
    \centering
    \includegraphics[width=\linewidth,keepaspectratio]{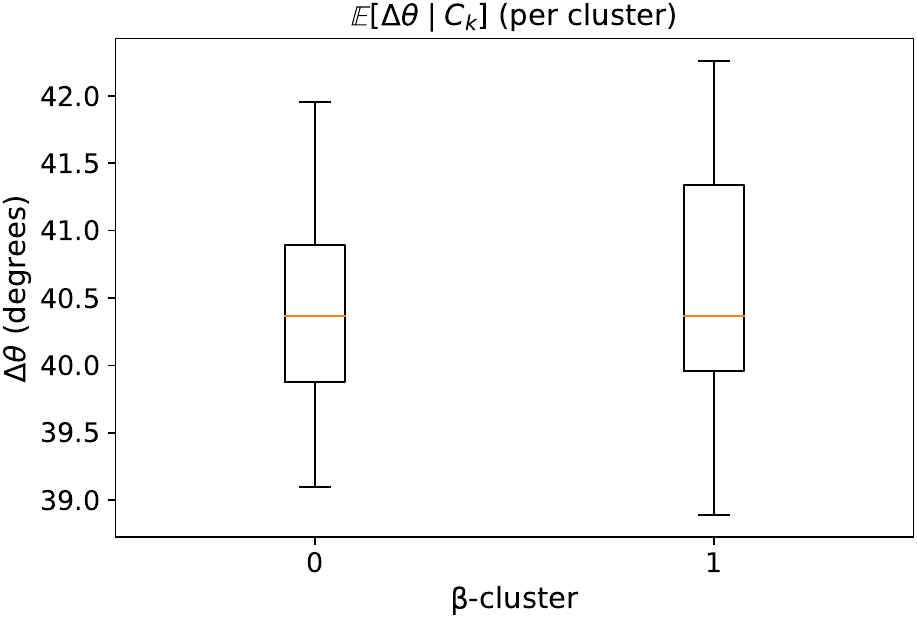}
    \caption{$\mathbb{E}[\Delta\theta \mid C_k]$ by cluster.}
  \end{subfigure}\hfill
  \begin{subfigure}[t]{0.32\linewidth}
    \centering
    \includegraphics[width=\linewidth,keepaspectratio]{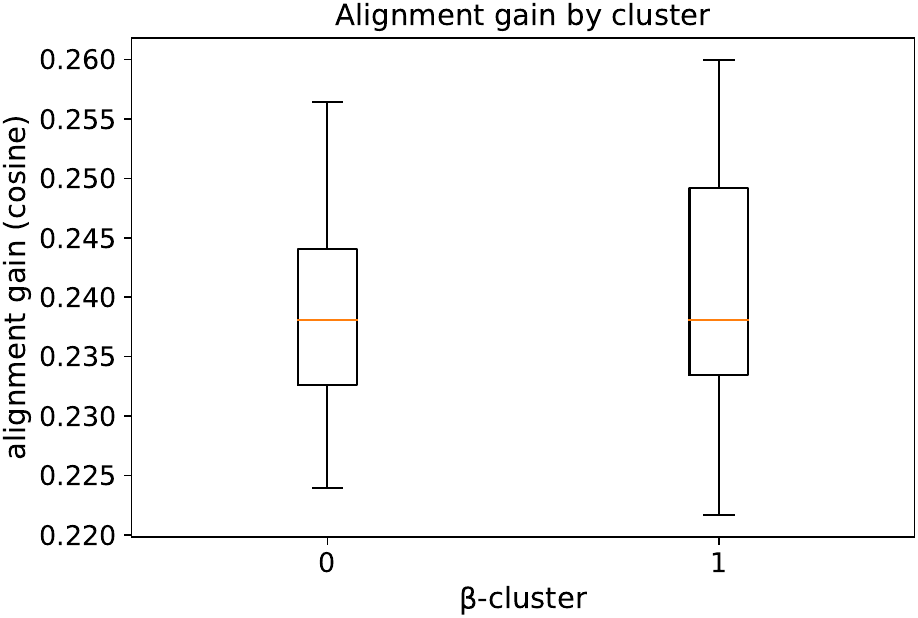}
    \caption{Alignment gain by cluster.}
  \end{subfigure}\hfill
  \begin{subfigure}[t]{0.32\linewidth}
    \centering
    \includegraphics[width=\linewidth,keepaspectratio]{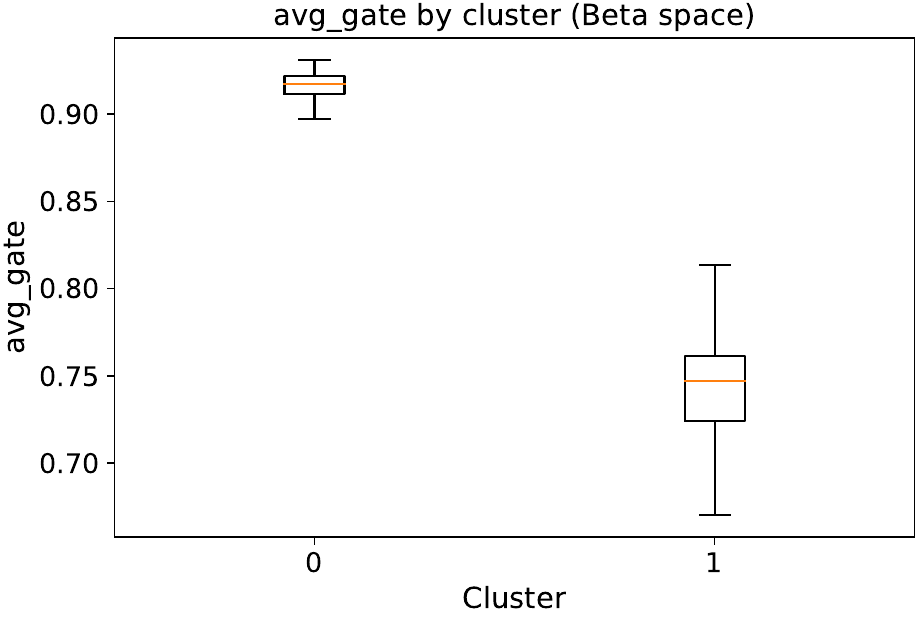}
    \caption{$\overline{\mathrm{gate}}$ by cluster (sanity).}
  \end{subfigure}

  \caption{Directional evidence with reflowed layout. (a)~Consistent $\mathbf{v}_{\text{text}}\!\rightarrow\!\mathbf{v}_{\text{img+txt}}$ rotation; (b)~positive alignment gain to $\mathbf{v}^{\text{steer}}$; (c,d)~cluster-wise stability; (e)~systematic gate differences.}
  \label{fig:vis_angles}
\end{figure}

\paragraph{Quantifying directional structure.}
We quantify two effects central to our hypothesis:
\textbf{(i) Rotation from text-only to image-conditioned CARD:}
$\Delta\theta=\arccos\langle \mathbf{v}_{\text{text}}, \mathbf{v}_{\text{img+txt}}\rangle$ shows a tight distribution around $\sim$40$^\circ$ (mean $\approx$40.5$^\circ$, median $\approx$40.4$^\circ$), indicating a consistent, non-trivial visual-induced rotation rather than random drift (Fig.~\ref{fig:vis_angles}a); this effect persists across steering clusters (Fig.~\ref{fig:vis_angles}c).
\textbf{(ii) Alignment gain w.r.t.\ steering:}
$\langle \mathbf{v}_{\text{img+txt}}, \mathbf{v}^{\text{steer}}\rangle - \langle \mathbf{v}_{\text{text}}, \mathbf{v}^{\text{steer}}\rangle$ is positive on average (mean $\approx$0.239, median $\approx$0.238) and remains positive across clusters (Figs.~\ref{fig:vis_angles}b,d), showing that image-conditioned $\mathbf{v}_{\text{CARD}}$ is \emph{closer} to the actual steering geometry used by the $\beta$-gate. Together, these results substantiate our motivation: aggregating self-attention residual updates yields a robust sample-specific visual-evidence direction that aligns with the downstream steering mechanism.

% ===== Reflowed layout: row of two histograms (full width), row of three cluster plots (full width) =====

\paragraph{Notes.}
Silhouette scores are typically higher for $\mathbf{v}_t^{\text{steer}}$ than for $\mathbf{v}_{\text{CARD}}$, consistent with the gate organizing/scaling directions across samples.
We emphasize that t\hbox{-}SNE primarily supports \emph{local} neighborhood interpretation; all scalar statistics are computed in the original vector spaces.

\begin{figure*}[t]
  \centering
  \includegraphics[width=0.86\textwidth]{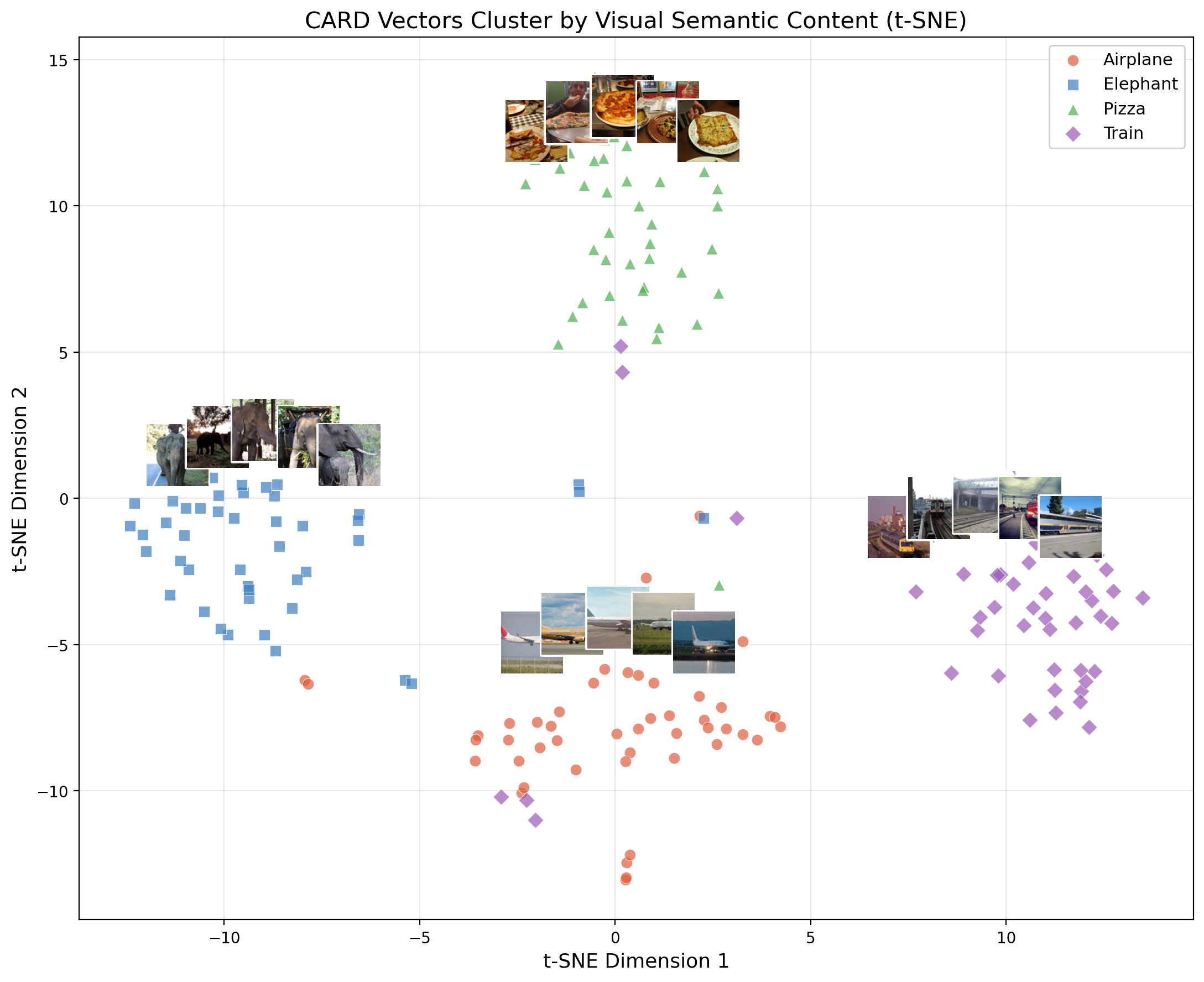}
  \caption{\textbf{CARD vectors cluster by visual semantic content.} We compute CARD vectors from LLaVA-1.5-7B on four COCO categories with 50 images per category and visualize them using t-SNE. The category-level clustering indicates that CARD captures image-conditioned visual semantics rather than random activation noise.}
  \label{fig:card_semantic_tsne}
\end{figure*}

\paragraph{Semantic clustering of CARD.}
In addition to the paired geometry above, we directly test whether CARD encodes category-level visual semantics. We extract CARD vectors from LLaVA-1.5-7B on four COCO categories with 50 images per category and visualize them with t-SNE. Figure~\ref{fig:card_semantic_tsne} shows that samples from the same visual category form coherent neighborhoods, suggesting that CARD captures image-conditioned semantic structure rather than random activation noise.

\section{Additional Experiment}

\subsection{LLaVA Internal Dynamic Analysis}
\label{sec:llava_dynamics}

\begin{figure}[t]  % !tH强制置顶当前页，确保2个内容块紧凑共页
  \centering
  \vspace*{0pt}  % 清除页顶残留间距，紧贴顶部
  \setlength{\abovecaptionskip}{0pt}
  \setlength{\belowcaptionskip}{4pt}

  % 第一部分：card analysis的3个子图（纵向排列，压缩间距）
  \begin{subfigure}[t]{\textwidth}
    \centering
    \includegraphics[width=\linewidth]{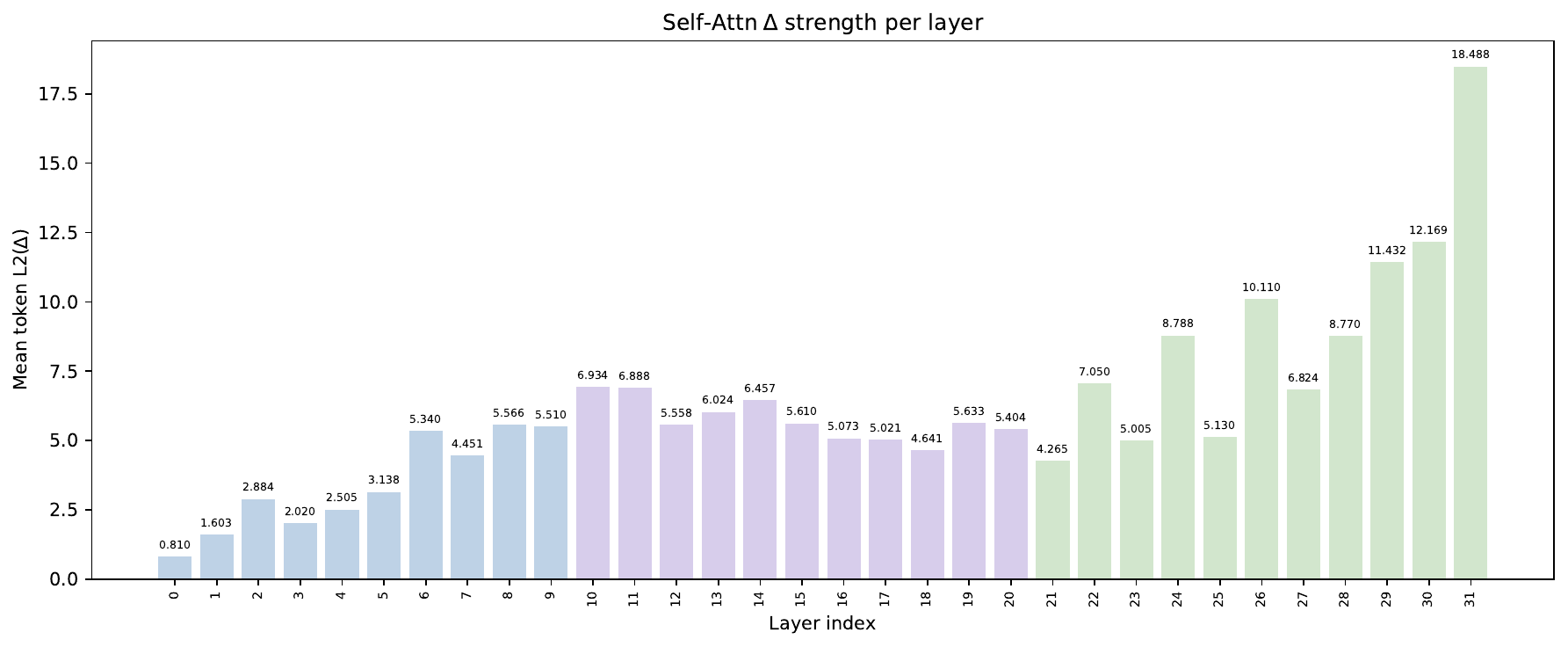}
    \caption{Absolute strength of the update vector ($\|\delta^l\|$) across layers.}
    \label{fig:delta_norm}
    \vspace{2pt}  % 缩小子图间间距，为case study腾出空间
  \end{subfigure}
  
  \begin{subfigure}[t]{\textwidth}
    \centering
    \includegraphics[width=\linewidth]{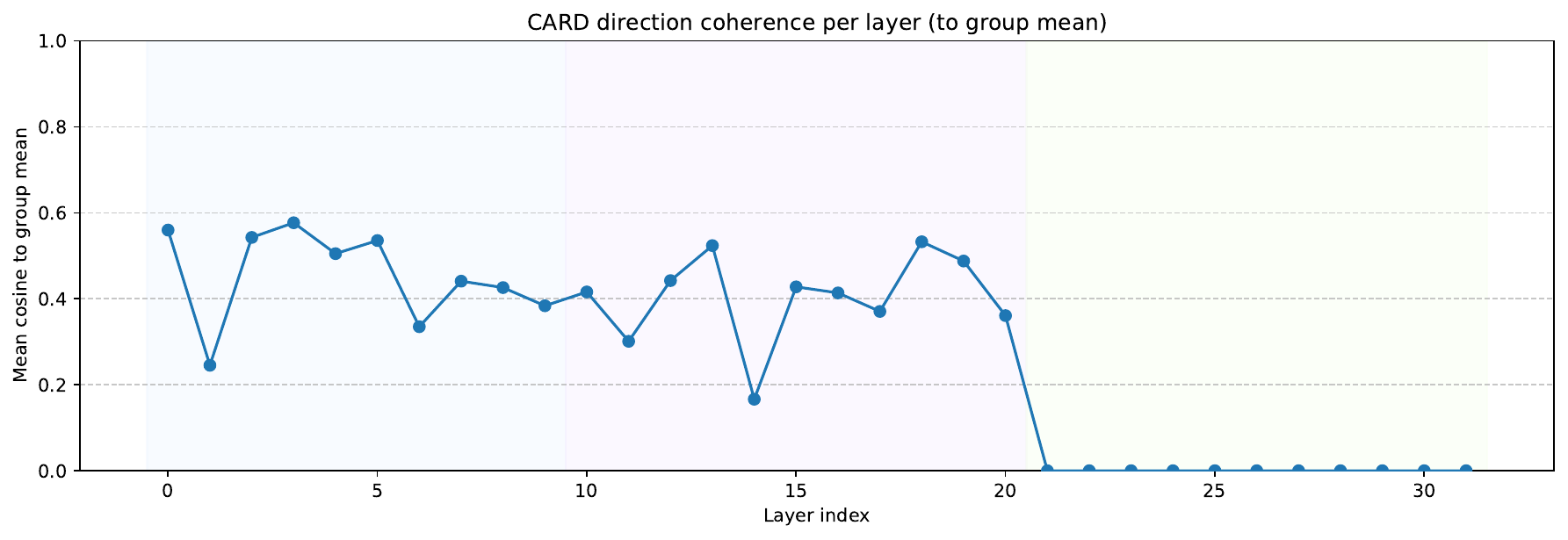}
    \caption{Directional coherence within layer groups.}
    \label{fig:alignment}
    \vspace{2pt}
  \end{subfigure}
  
  \begin{subfigure}[t]{\textwidth}
    \centering
    \includegraphics[width=\linewidth]{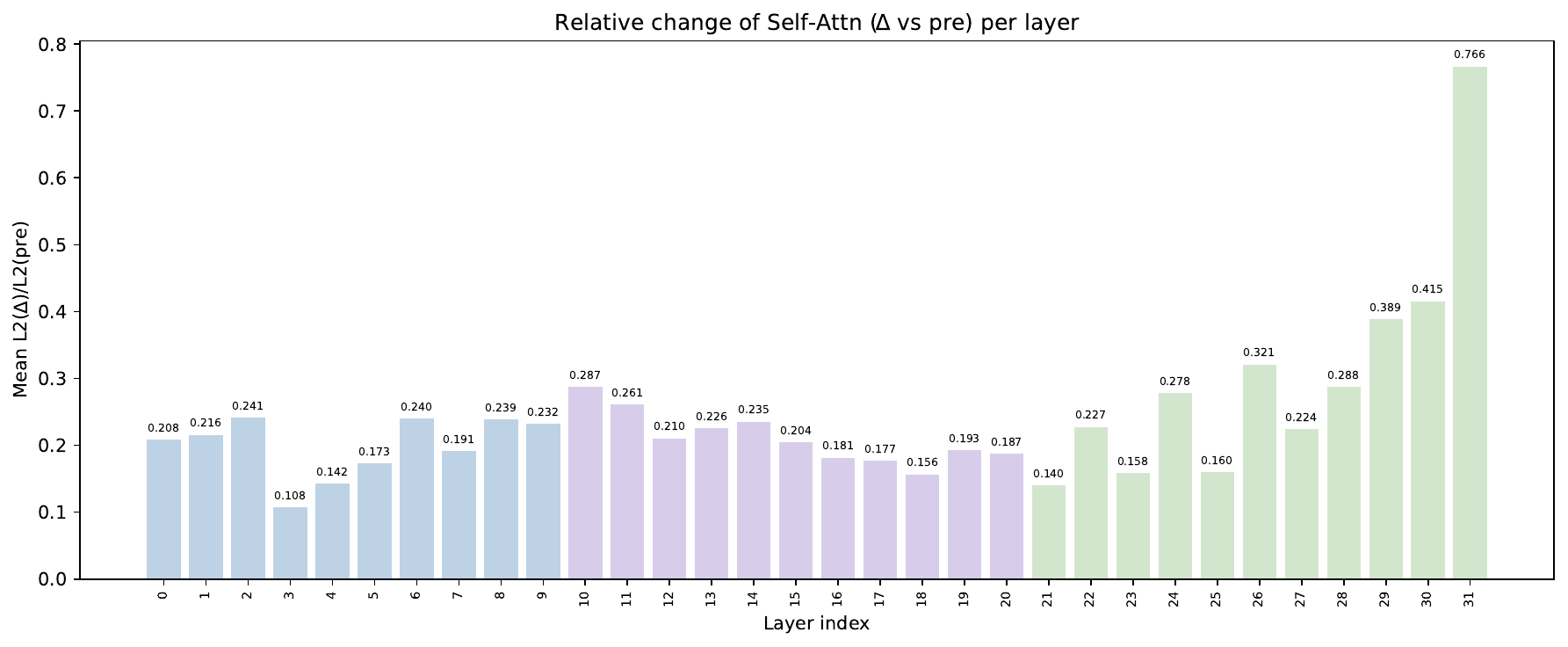}
    \caption{Relative strength of the update vector ($\|\delta^l\| / \|h_{\text{pre}}^l\|$).}
    \label{fig:ratio}
    \vspace{6pt}  
  \end{subfigure}
  \caption{Analysis of the internal dynamics of LLaVA-1.5 \citep{liu_improved_2024}. (a, c) Both the absolute and relative strength of self-attention updates peak in the middle-to-late layers, which identify a "computational core". (b) Mid-late layers show significantly lower directional coherence than other regions.}
  \label{fig:card_analysis_method}
\end{figure}

As mentioned in Section 3.2, our method is guided by an analysis of the internal dynamics of LLaVA-1.5~\citep{liu_improved_2024}, with key findings illustrated in Figure~\ref{fig:card_analysis_method}. By examining the residual update vector from the self-attention module at each layer, we identified two properties that informed our intervention strategy: 

\textbf{Intervention Leverage Peaks in Late Layers.} We can find that the magnitude of the residual update, which represents the ``leverage'' an intervention can have, is not uniform. We found that its strength grows with model depth, peaking in the late decoder blocks (approx. layers 26-32). This indicates that interventions in these layers have the greatest potential to influence the model's final output (Figure~\ref{fig:delta_norm}, \ref{fig:ratio}). 

\textbf{Directional Coherence Collapses in Late Layers.} 
While late layers offer the most leverage, the directional coherence of their update vectors collapses after approximately layer 21 (Figure~\ref{fig:alignment}). Coherence is moderate only in the early-to-mid layers. This suggests that applying a fixed, global steering vector in the high-leverage late layers is suboptimal, as the intervention may be misaligned with the model's unstable internal state.

\subsection{Additional Ablation Results}
\label{sec:additional_ablation}
\begin{figure*}[t] % 注意这里加了 *
  \begin{subfigure}[t]{0.33\linewidth} 
    \centering
    \includegraphics[width=\linewidth,keepaspectratio]{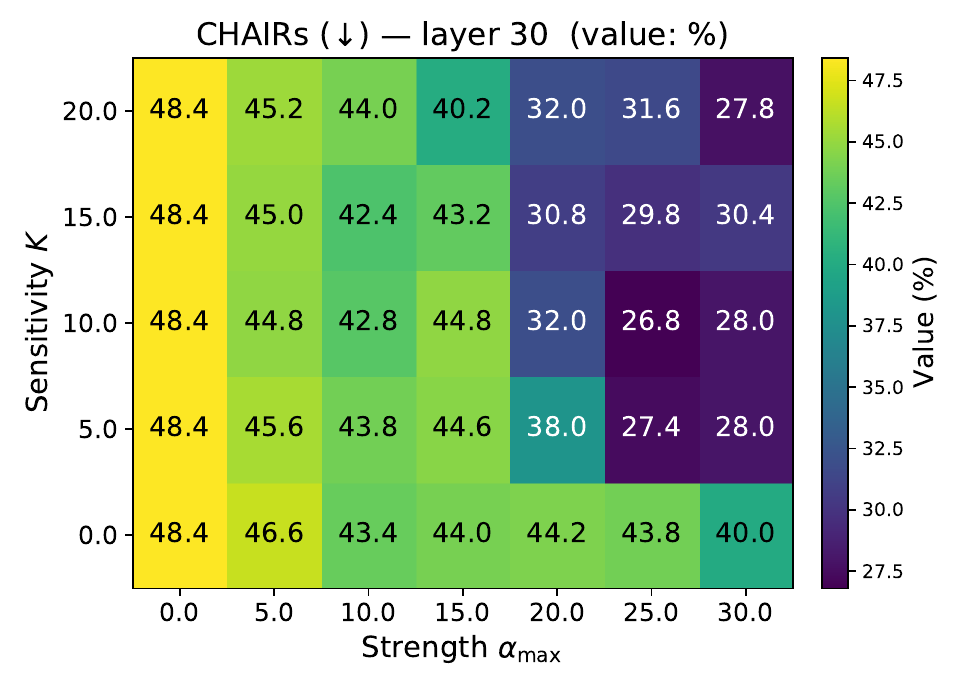}
    \caption{CHAIR$_\text{S}$ $\alpha_\text{max}\times k$ heatmap.}
  \end{subfigure}\hfill
  \begin{subfigure}[t]{0.33\linewidth}
    \centering
    \includegraphics[width=\linewidth,keepaspectratio]{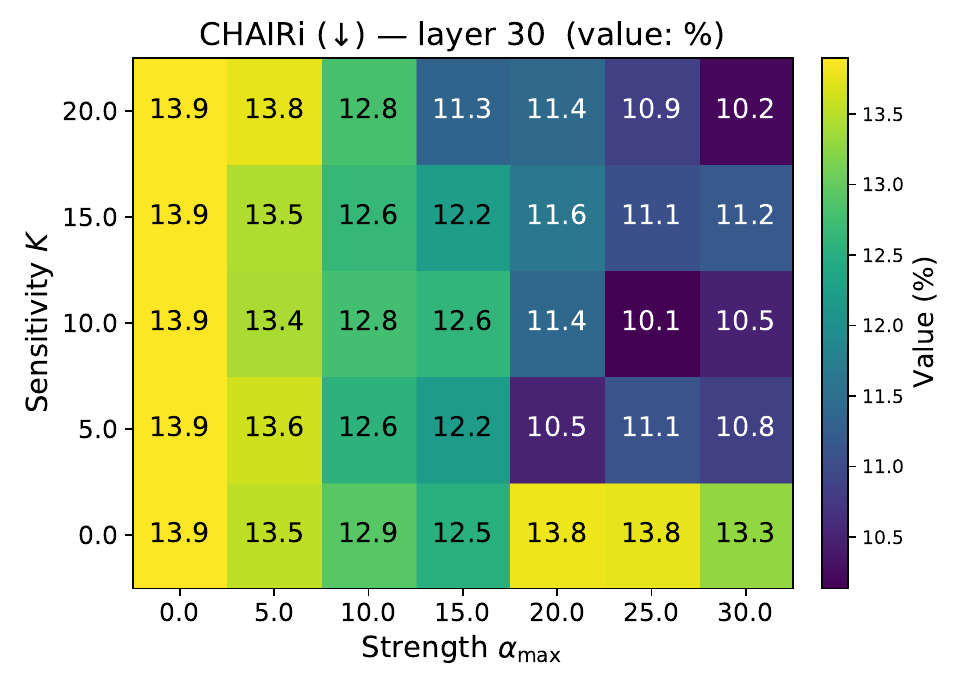}
    \caption{CHAIR$_\text{i}$ $\alpha_\text{max}\times k$ heatmap. }
  \end{subfigure}\hfill
  \begin{subfigure}[t]{0.33\linewidth}
    \centering
    \includegraphics[width=\linewidth,keepaspectratio]{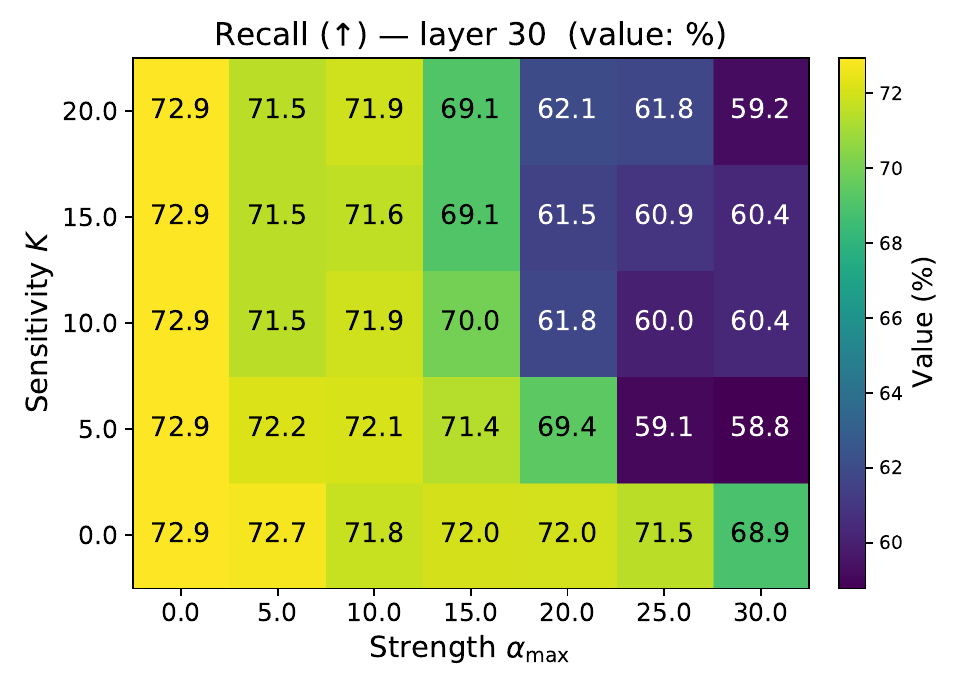}
    \caption{$Recall$ $\alpha_\text{max}\times k$ heatmap.}
  \end{subfigure}
  \caption{Ablation matrices for RUDDER on LLaVA-1.5~\citep{liu_improved_2024}}
  \label{fig:ablation_llava}
\end{figure*}

\begin{figure*}[t]
  \begin{subfigure}[t]{0.33\linewidth} 
    \centering
    \includegraphics[width=\linewidth,keepaspectratio]{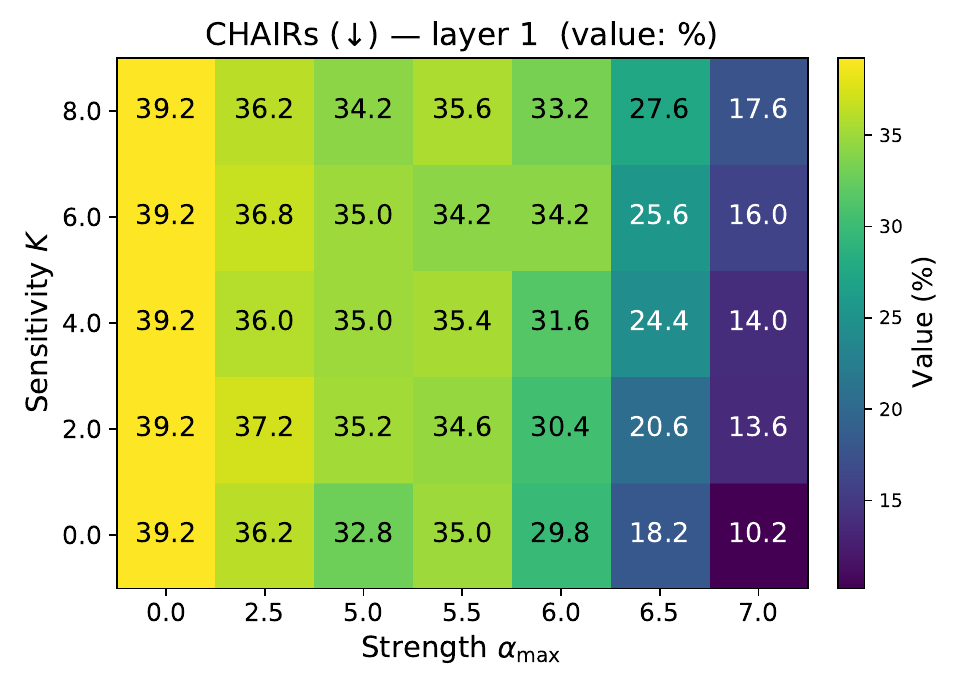}
    \caption{CHAIR$_\text{S}$ $\alpha_\text{max}\times k$ heatmap.}
  \end{subfigure}\hfill 
  \begin{subfigure}[t]{0.33\linewidth}
    \centering
    \includegraphics[width=\linewidth,keepaspectratio]{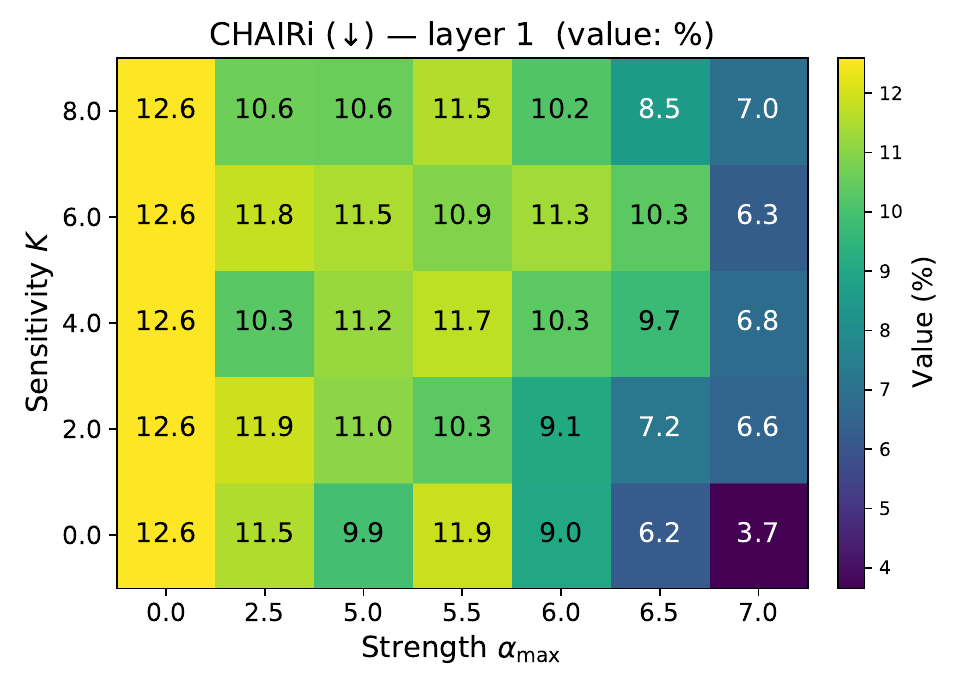}
    \caption{CHAIR$_\text{i}$ $\alpha_\text{max}\times k$ heatmap. }
  \end{subfigure}\hfill
  \begin{subfigure}[t]{0.33\linewidth}
    \centering
    \includegraphics[width=\linewidth,keepaspectratio]{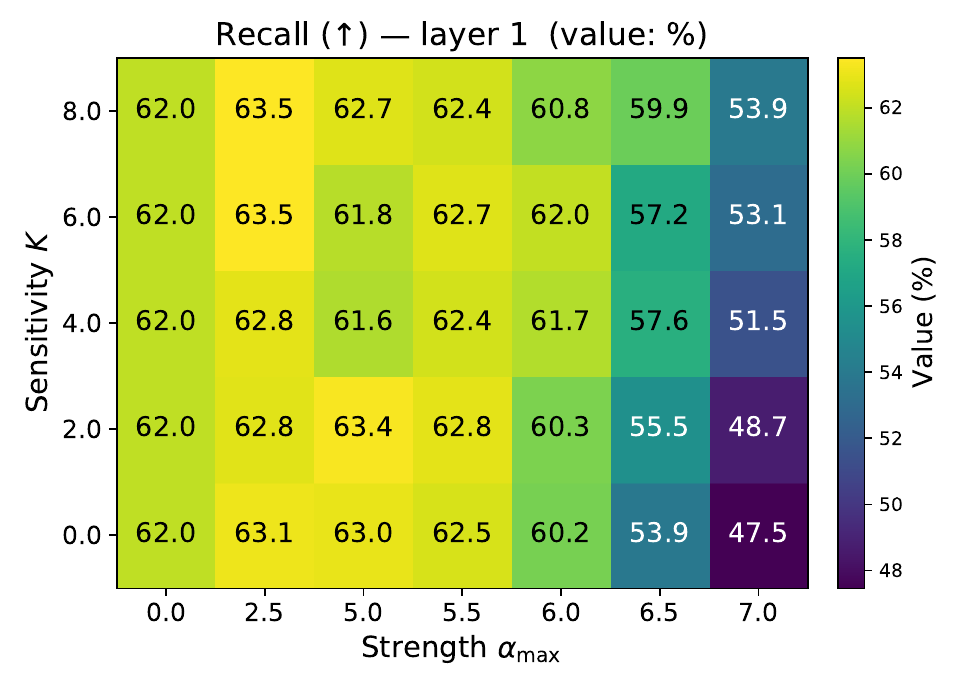}
    \caption{$Recall$ $\alpha_\text{max}\times k$ heatmap.}
  \end{subfigure}
  \caption{Ablation matrices for RUDDER on InstructBLIP~\citep{dai_instructblip_2023}}
  \label{fig:ablation_InstructBlip}
\end{figure*}

We provide supplementary abalation results on LLaVA-1.5 and InstructBLIP, as shown in Figure~\ref{fig:ablation_llava} and Figure~\ref{fig:ablation_InstructBlip}.
These analyses complement the main ablation study conducted on Idefics2 in Section~\ref{sec:ablation}.

\begin{figure*}[t]
  \centering
  \includegraphics[width=0.88\textwidth]{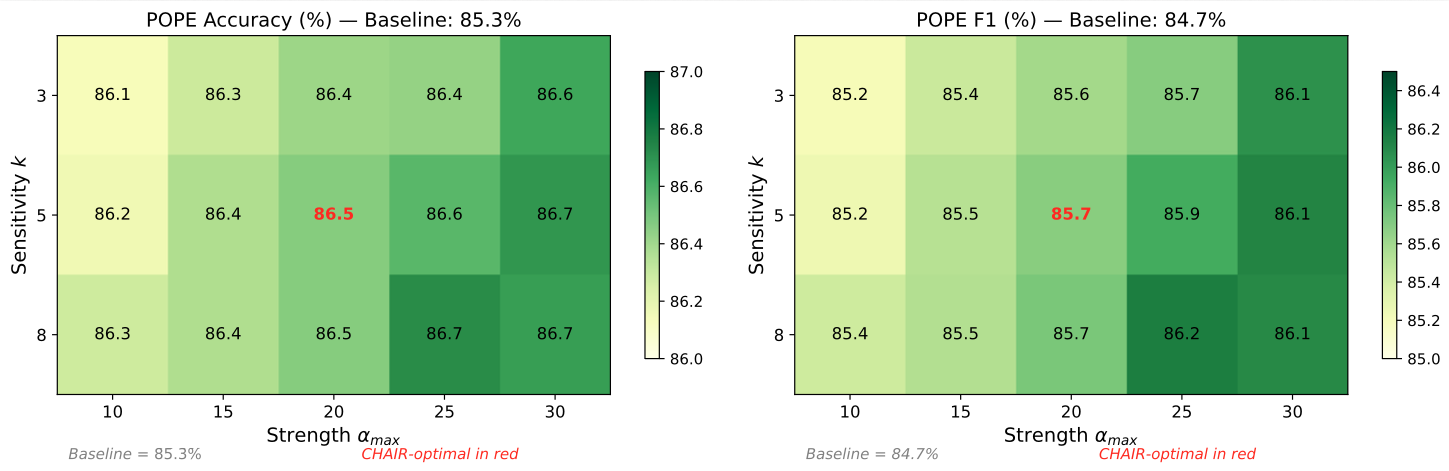}
  \caption{\textbf{POPE hyperparameter sensitivity on LLaVA-1.5-7B.} All tested configurations improve over the baseline, with a small spread across the grid. The CHAIR-tuned setting remains close to the best POPE configuration, suggesting that RUDDER is not highly sensitive to the exact choice of $\alpha_{\max}$ and $k$.}
  \label{fig:pope_sensitivity}
\end{figure*}

We further verify whether the CHAIR-tuned hyperparameters transfer to POPE. Figure~\ref{fig:pope_sensitivity} shows that all tested configurations improve over the vanilla baseline, and the CHAIR-selected setting remains close to the best POPE configuration. This supports the use of a small held-out calibration set rather than benchmark-specific retuning.

The results for both models confirm the same core trends observed with Idefics2. Specifically, the heatmaps reveal a consistent trade-off between hallucination mitigation and recall. As the steering strength $\alpha_{max}$ increases, both CHAIRS and CHAIRI scores improve (decrease), but this is often accompanied by a drop in recall. The gate sensitivity parameter, $k$, plays a similar, non-linear modulating role in this balance. While the general trade-off is consistent, the optimal hyperparameter values vary by model architecture, highlighting the need for model-specific tuning.

\vspace{-4pt}
% ------------------- 第2页：3个连续case study（独占1页，位于页面上方） -------------------
\subsection{Case Study}
\label{sec:case_study}
\begin{figure*}[t] % 注意这里加了 *
  \centering
  
  % --- 子图1: Case Study 1 ---
  \begin{subfigure}{\linewidth}
      \centering
      \includegraphics[width=\linewidth, keepaspectratio]{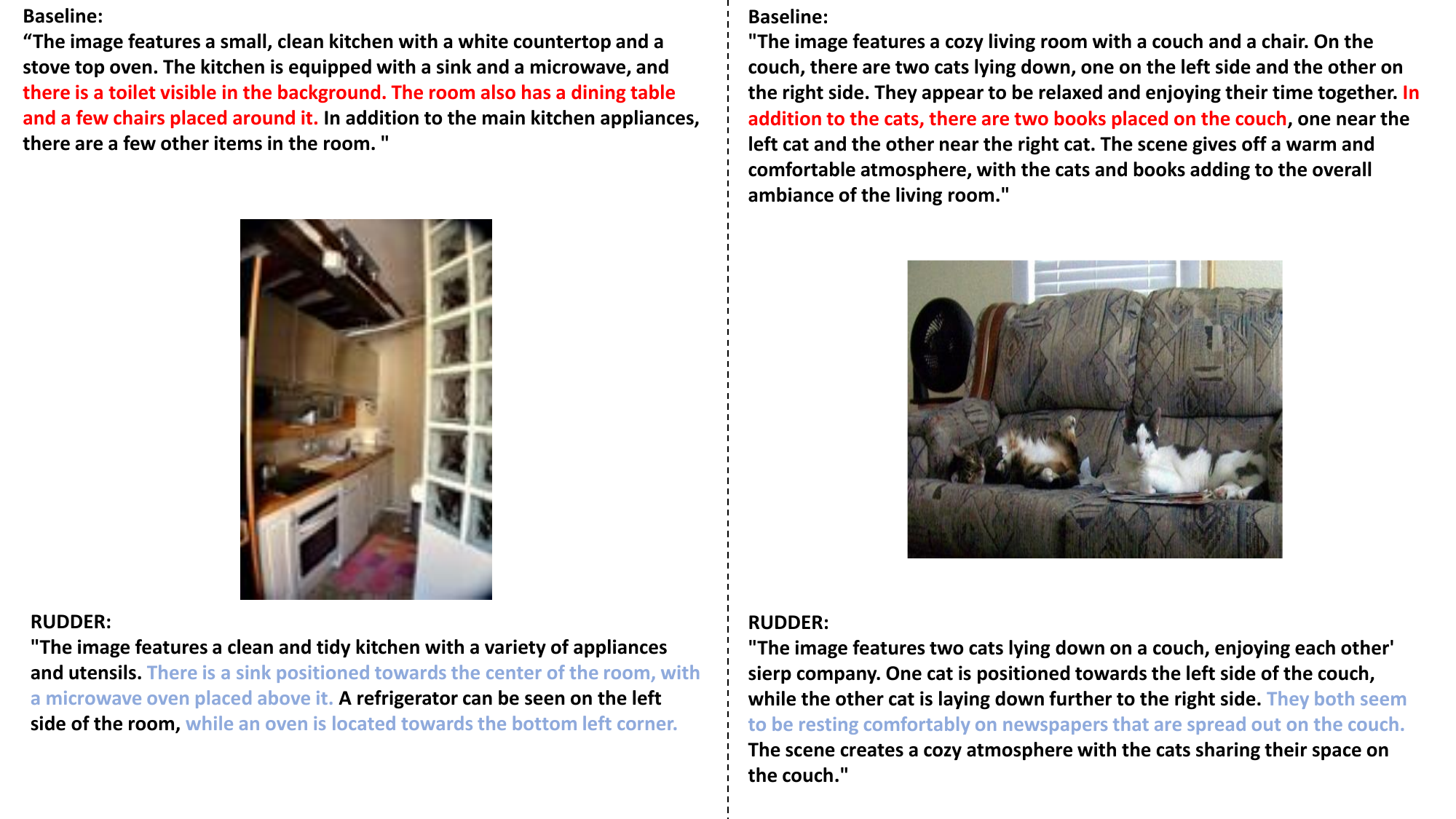}
      \label{fig:case1}
  \end{subfigure}
  
  \vspace{4pt} % 间距

  % --- 子图2: Case Study 2 ---
  \begin{subfigure}{\linewidth}
      \centering
      \includegraphics[width=\linewidth, keepaspectratio]{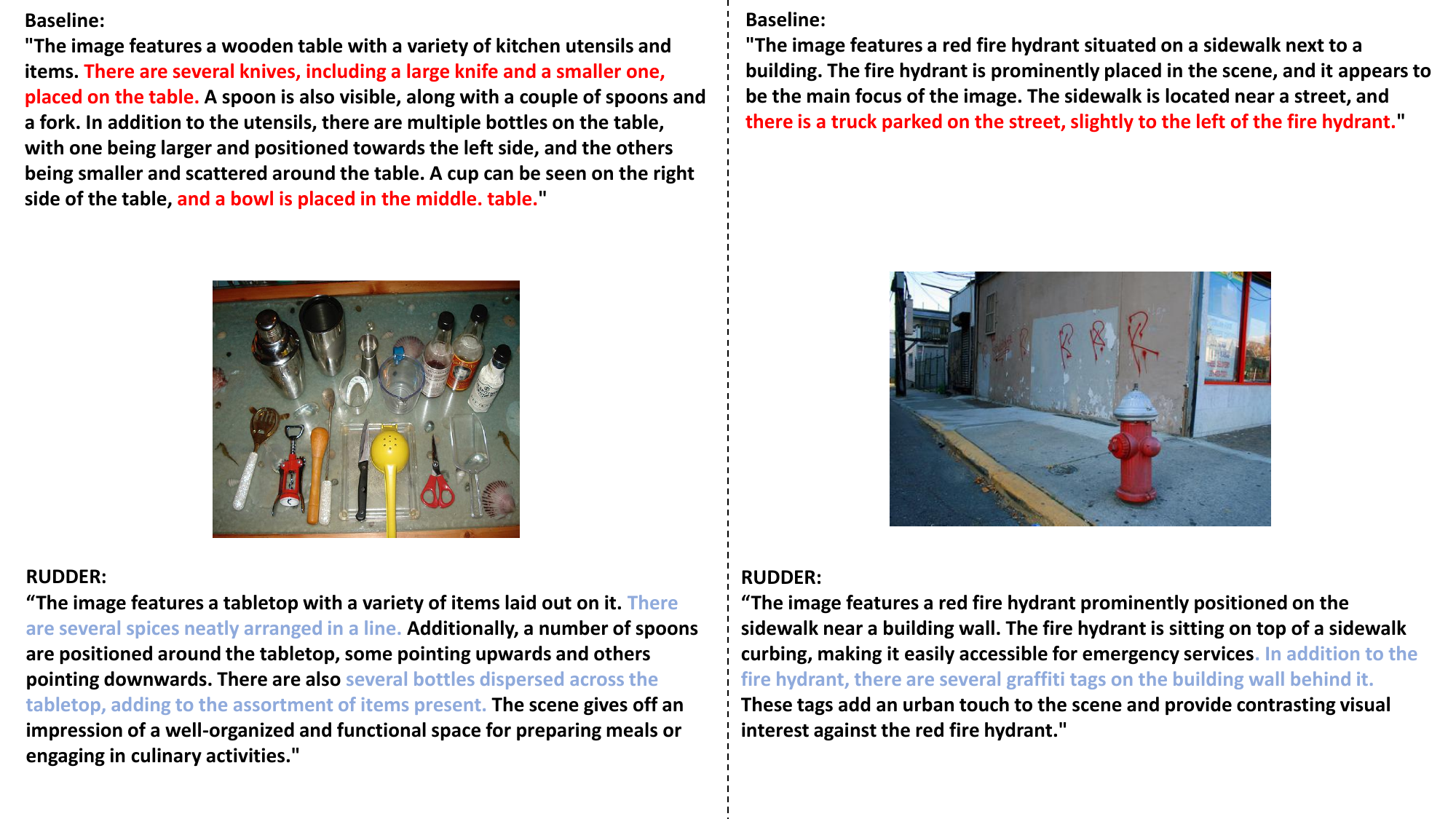}
      \label{fig:case2}
  \end{subfigure}
  
  \caption{Case study. Hallucinated contents generated by the vanilla LLaVA-1.5 are marked in \textcolor{red}{red}, while the factual contents from our method are colored with \textcolor{cyan}{blue}.}
  \label{fig:cases_1_and_2}
\end{figure*}

\begin{figure*}[t]
  \centering

  % --- 子图3: Case Study 3 ---
  \begin{subfigure}{\linewidth}
      \centering
      \includegraphics[width=\linewidth, keepaspectratio]{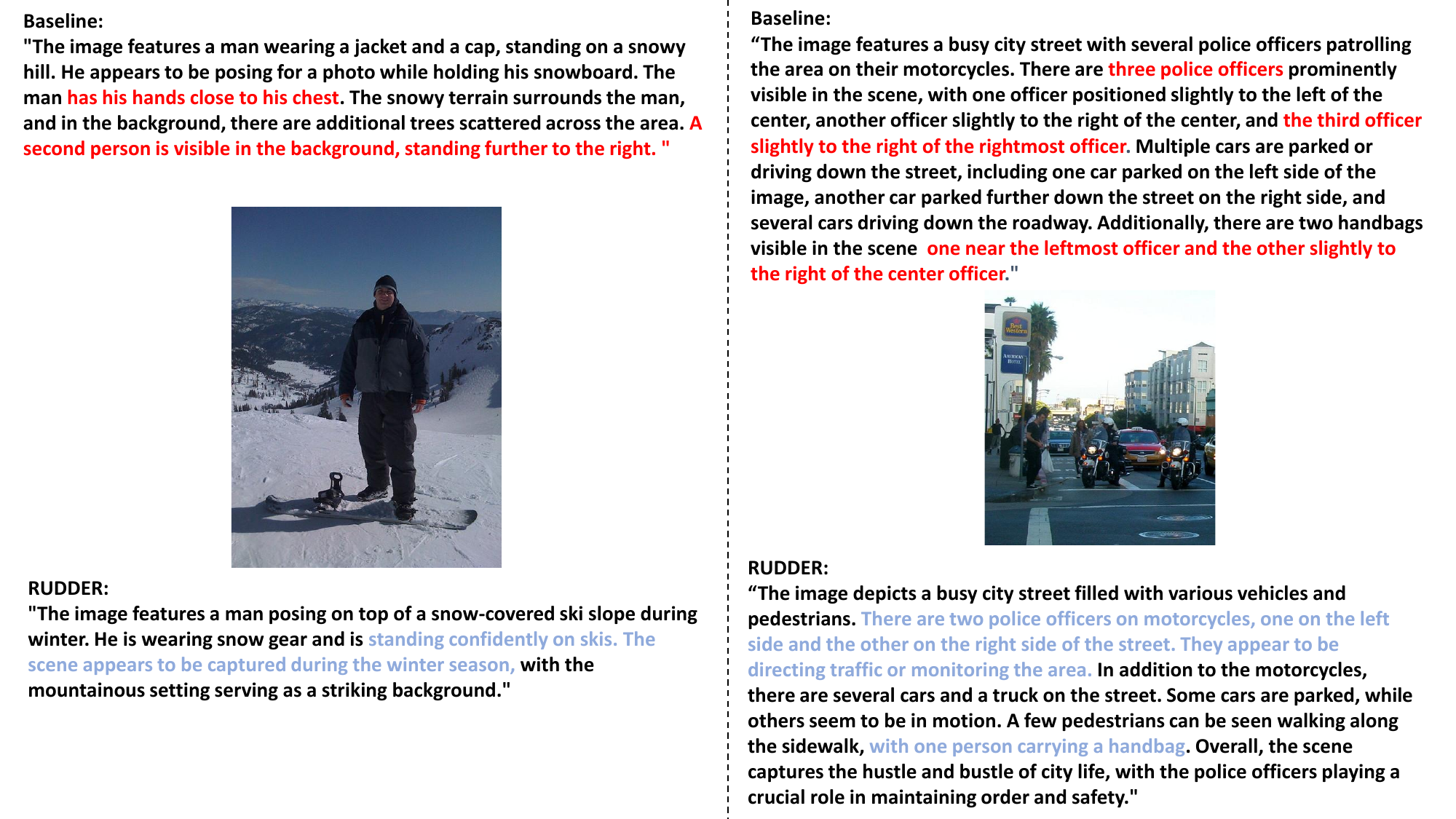}
      \label{fig:case3}
  \end{subfigure}
  
  \vspace{4pt}

  % --- 子图4: Case Study 4 ---
  \begin{subfigure}{\linewidth}
      \centering
      \includegraphics[width=\linewidth, keepaspectratio]{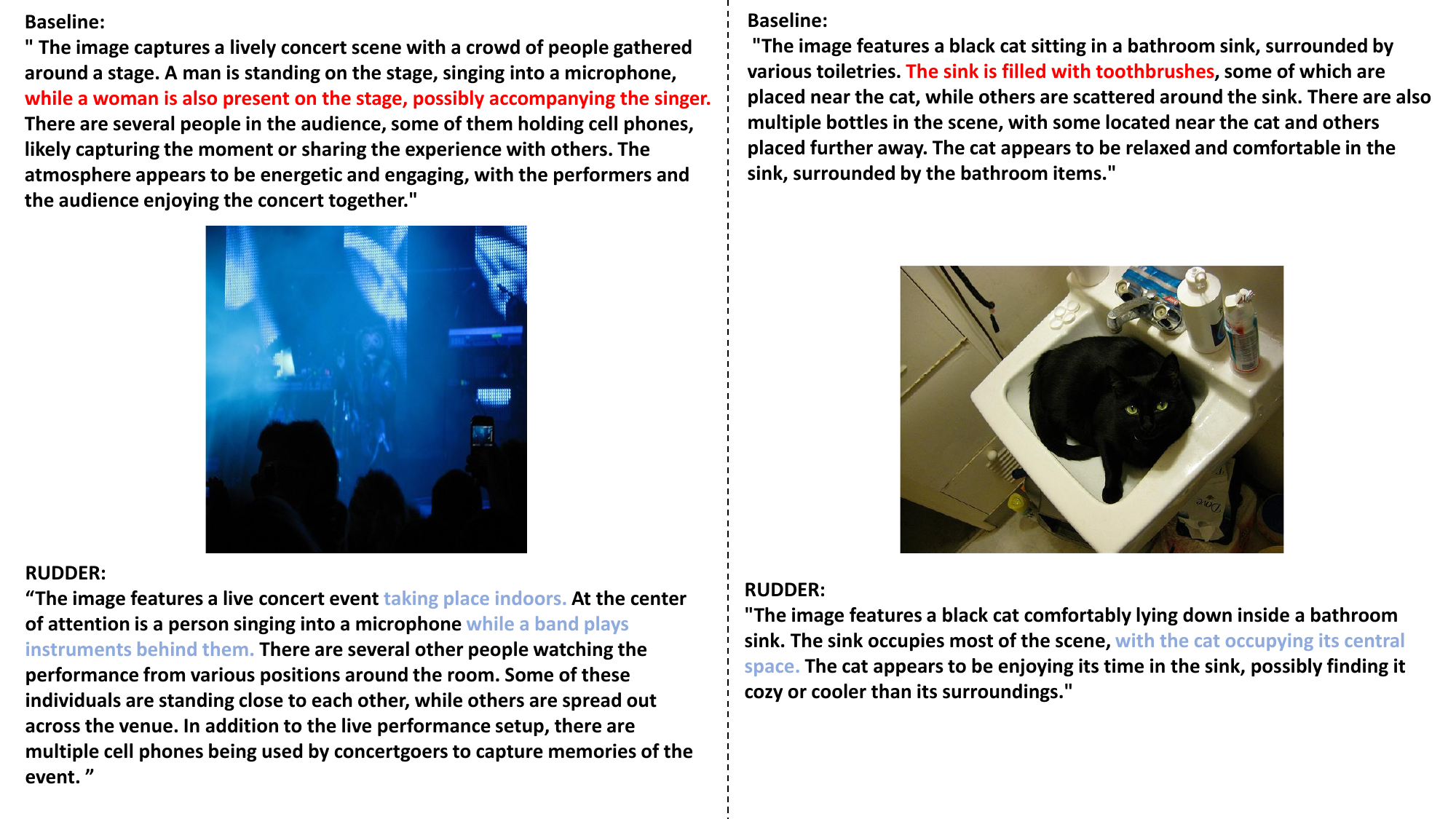}
      \label{fig:case4}
  \end{subfigure}
  
  \caption{Case study. Hallucinated contents generated by the vanilla Idefics2 are marked in \textcolor{red}{red}, while the factual contents from our method are colored with \textcolor{cyan}{blue}.}
  \label{fig:cases_3_and_4}
\end{figure*}

\begin{figure*}[t]
  \centering
  
  % --- 子图1: Case Study 1 ---
  \begin{subfigure}{\linewidth}
      \centering
      \includegraphics[width=\linewidth, keepaspectratio]{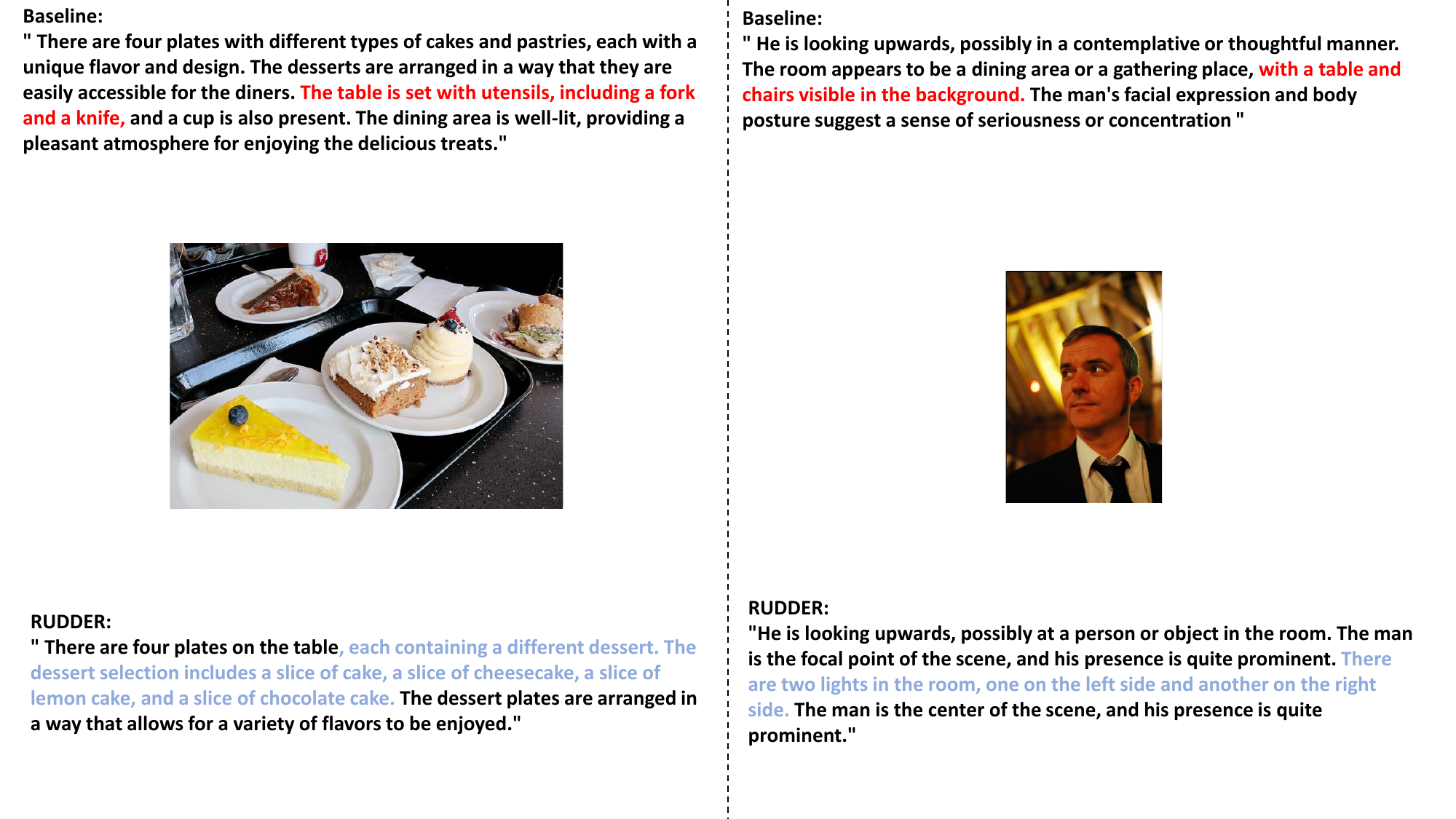}
      \label{fig:case6}
  \end{subfigure}
  
  \vspace{4pt}

  % --- 子图2: Case Study 2 ---
  \begin{subfigure}{\linewidth}
      \centering
      \includegraphics[width=\linewidth, keepaspectratio]{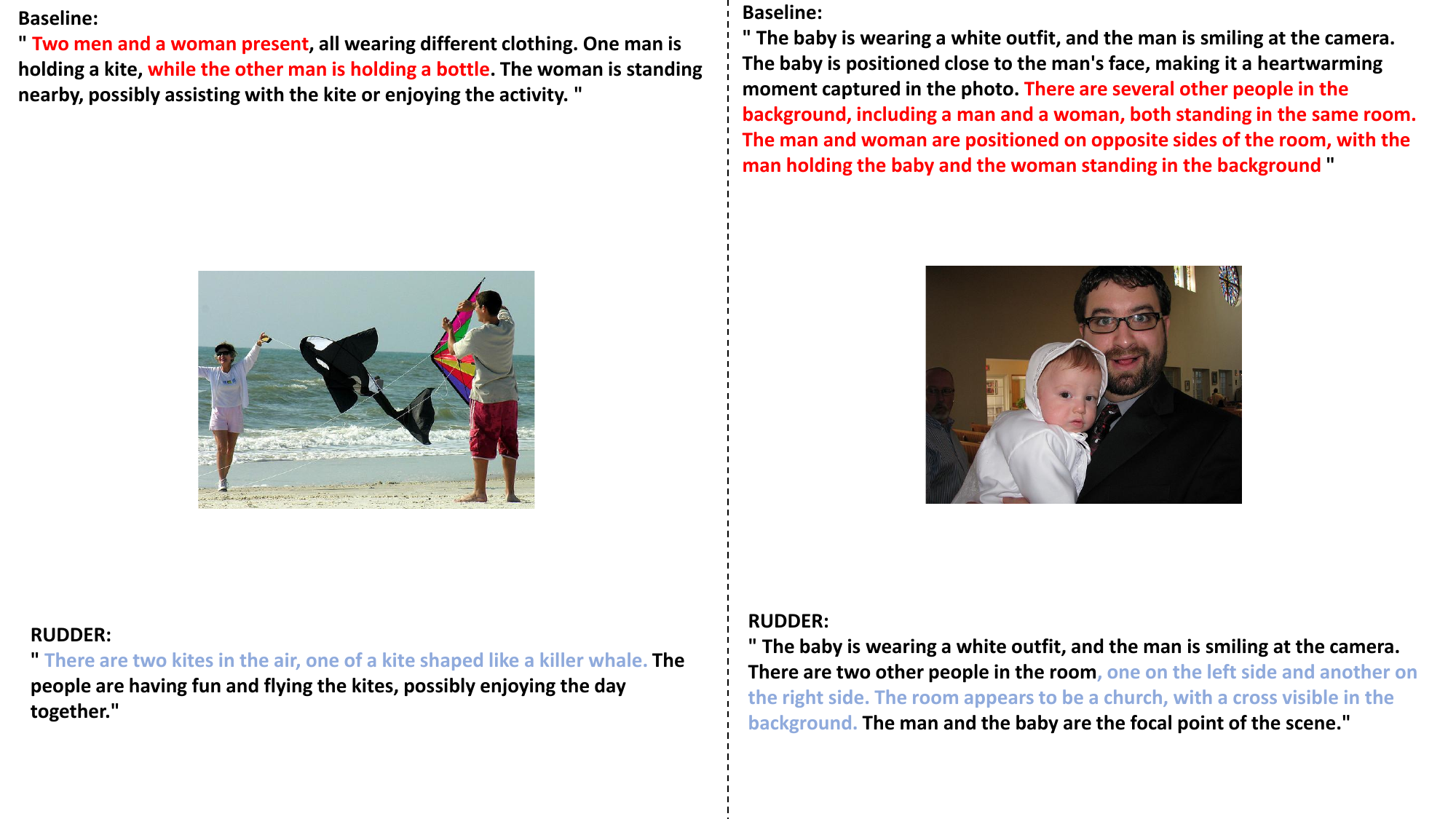}
      \label{fig:case7}
  \end{subfigure}
  
  \caption{Case study. Hallucinated contents generated by the vanilla InstructBLIP are marked in \textcolor{red}{red}, while the factual contents from our method are colored with \textcolor{cyan}{blue}.}
  \label{fig:cases_6_and_7}
\end{figure*}

To provide a qualitative illustration of our method's real-world performance, we present a series of case studies in Figures~\ref{fig:cases_1_and_2}, ~\ref{fig:cases_3_and_4} and ~\ref{fig:cases_6_and_7}. 
From these examples, we can find that RUDDER is highly effective at eliminating object-level hallucinations. It successfully removes entirely non-existent objects from the captions (e.g., a hallucinated ``second person" or ``cup"), and corrects misidentified objects (e.g., correctly identifying ``skis" instead of ``snowboard"). Moreover, RUDDER's outputs are not only more factually accurate but also more semantically cautious. The corrected descriptions often adopt more conservative language, such as using phrases like ``appears to be", ``may be", or ``suggesting that". By replacing the baseline models' confident yet incorrect assertions with more grounded and appropriately qualified statements, RUDDER significantly enhances the overall reliability and trustworthiness of the generated text.

\begin{figure*}[t]
  \centering
  \begin{subfigure}[t]{0.28\textwidth}
    \centering
    \includegraphics[width=\linewidth]{figures//COCO_val2014_000000309510.jpg}
    \caption{Input image.}
  \end{subfigure}\hfill
  \begin{subfigure}[t]{0.69\textwidth}
    \centering
    \includegraphics[width=\linewidth]{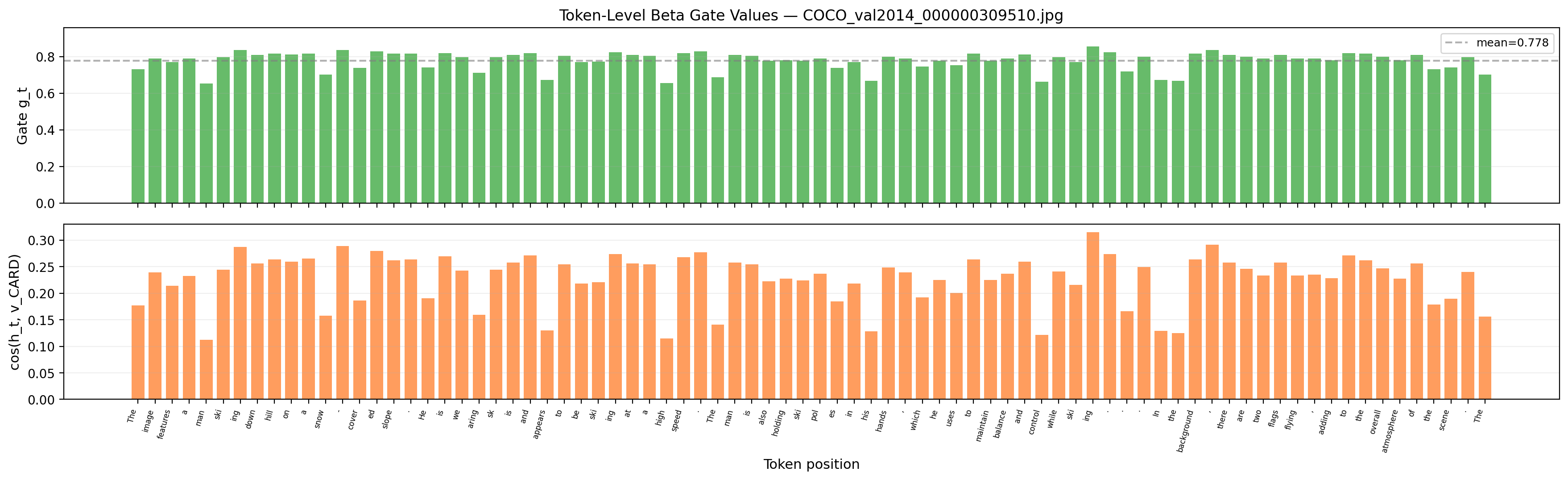}
    \caption{Token-level gate and CARD similarity.}
  \end{subfigure}
  \caption{\textbf{Token-level Beta Gate visualization for a skiing image.} Visually grounded tokens receive consistently high gate values, while less visually grounded or more abstract tokens show lower CARD similarity. This supports the interpretation of the gate as a visual-grounding-dependent trust signal.}
  \label{fig:gate_ski_case}
\end{figure*}

\begin{figure*}[t]
  \centering
  \begin{subfigure}[t]{0.28\textwidth}
    \centering
    \includegraphics[width=\linewidth]{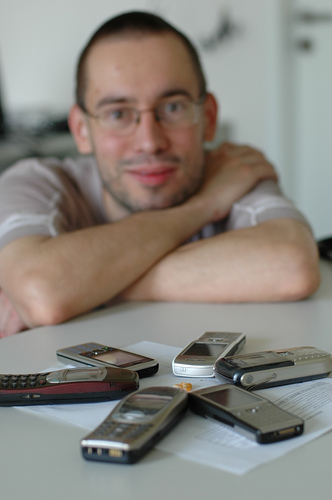}
    \caption{Input image.}
  \end{subfigure}\hfill
  \begin{subfigure}[t]{0.69\textwidth}
    \centering
    \includegraphics[width=\linewidth]{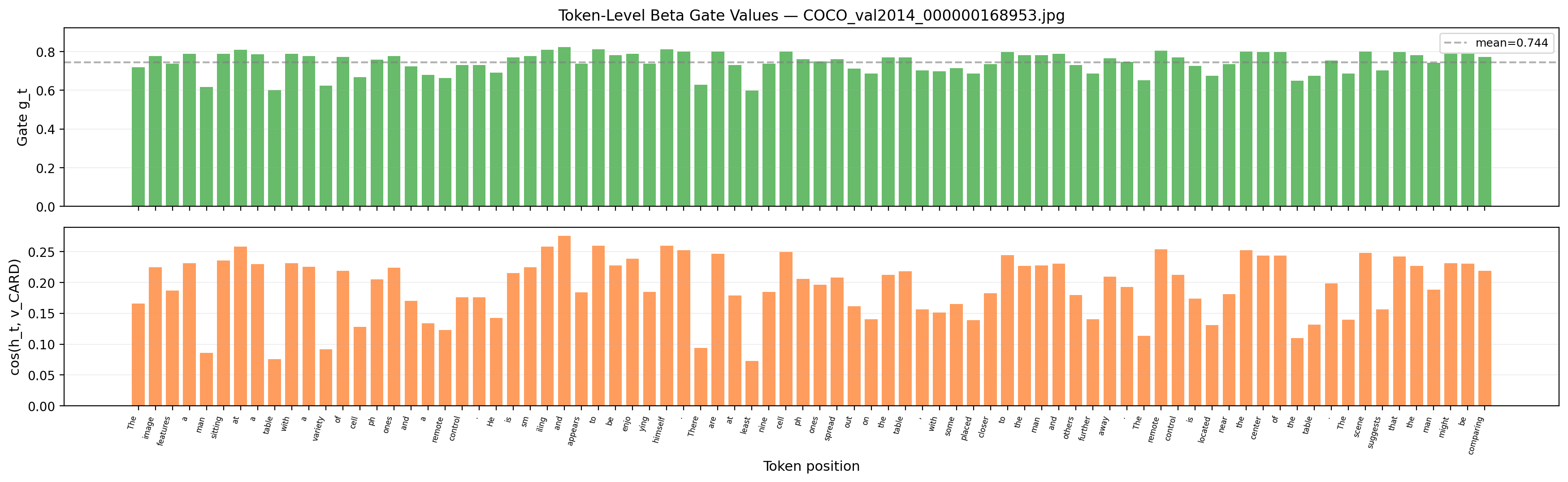}
    \caption{Token-level gate and CARD similarity.}
  \end{subfigure}
  \caption{\textbf{Token-level Beta Gate visualization for a cell-phone image.} The hallucinated token sequence related to ``remote control'' receives low CARD similarity, showing that the gate can down-weight intervention when the current hidden state is poorly aligned with the image-conditioned CARD direction.}
  \label{fig:gate_phone_case}
\end{figure*}

To further inspect the behavior of the adaptive gate, we visualize token-level gate values and CARD similarities for two representative samples. Figures~\ref{fig:gate_ski_case} and~\ref{fig:gate_phone_case} show that visually grounded content tokens generally receive stronger alignment and gate values, whereas tokens associated with unsupported concepts receive lower CARD similarity. These examples complement the caption-level case studies by showing that the Beta Gate operates at the intended token granularity.

% --- Case Study 1 & 2 (跨栏模式) ---

% --- Case Study 3 & 4 (跨栏模式) ---

% --- Case Study 6 & 7 (跨栏模式) ---

% --- LLaVA Heatmaps (跨栏模式) ---

% --- InstructBlip Heatmaps (跨栏模式) ---

\section{Clarifications Added After Reviewer Discussion}
\label{app:reviewer_clarifications}

\paragraph{Bayesian-inspired rather than full Bayesian inference.}
Our Beta Gate is not intended to perform full posterior inference over model parameters or generated tokens. Instead, it uses the Beta--Bernoulli posterior mean as a structured, bounded, and monotone mapping from visual-alignment evidence to an intervention strength. The softplus pseudo-counts ensure smoothness and positivity, the concentration parameter $c$ controls the prior strength, and the clamp range provides safety rails against degenerate gates.

\paragraph{Why reinforce high-alignment states.}
The gate should be interpreted as a trust estimator for the current visual trajectory, not as an error magnitude estimator. When $s_t=\cos(\mathbf{h}_{l,t},\mathbf{v}_{\mathrm{CARD}})$ is high, the hidden state is already aligned with the visual evidence extracted from prefill, so adding the CARD direction reinforces grounded generation. When $s_t$ is low, the token may be syntactic or the representation may be unstable; reducing the intervention avoids over-steering and helps preserve fluency and recall.

\paragraph{Layer selection across architectures.}
The best intervention layer is architecture-dependent. For LLaVA-style models and Idefics2, visual tokens are concatenated or projected into the LLM input, and the visual signal tends to fade in middle-to-late language layers; hence late layers offer stronger leverage. For Q-Former-based or stronger early-fusion models such as InstructBLIP and Qwen2.5-VL, useful multimodal alignment can occur earlier, so early intervention can be preferable. We therefore use a small held-out sweep to select a fixed layer per backbone and then keep that layer unchanged across benchmarks.

\paragraph{Efficiency claim.}
``No extra forward pass'' means that RUDDER does not invoke an additional model forward beyond the vanilla prefill and autoregressive decoding calls. CARD is extracted from the mandatory prefill pass through a read-only hook, while decoding only adds cheap vector operations and a residual-stream update. This is why RUDDER keeps the end-to-end latency overhead below $4\%$ in our measurements.

\paragraph{Hallucination--recall trade-off.}
A method can trivially reduce hallucination by becoming overly conservative. To avoid this failure mode, our calibration explicitly requires at least $95\%$ of baseline recall, and we additionally evaluate general multimodal capability on MME. This ensures that RUDDER reduces unsupported object mentions while largely preserving correct object recall and general visual reasoning behavior.

\section{LLMs Usage Statement}
Generative AI has been utilized to enhance the writing and to assist with coding tasks.
\end{document}